\pdfoutput=1
\documentclass[11pt]{article}
\usepackage[final]{acl}
\usepackage{times}
\usepackage{latexsym}
\usepackage[T1]{fontenc}
\usepackage[utf8]{inputenc}
\usepackage{microtype}
\usepackage{inconsolata}
\usepackage{amsmath}
\usepackage{amssymb}
\usepackage{xspace}
\usepackage{booktabs}
\usepackage{multirow}
\usepackage{graphicx}
\usepackage{tabularx}
\usepackage{subcaption}
\usepackage{tcolorbox}

\newcommand{\paratitle}[1]{\vspace{1.5ex}\noindent\textbf{#1}}
\newcommand{\ie}{\emph{i.e.,}\xspace}

\newcommand{\eg}{\emph{e.g.,}\xspace}

\newcommand{\ignore}[1]{}
\newcommand{\OURS}{{\textbf{GLoRE}}\xspace}

\title{Unlocking General Long Chain-of-Thought Reasoning Capabilities of Large Language Models via Representation Engineering}

\author{
    \textbf{
        Xinyu Tang\textsuperscript{\rm{1}\thanks{\ \ Equal contribution.}},
        Xiaolei Wang\textsuperscript{\rm{1}\footnotemark[1]},
        Zhihao Lv\textsuperscript{\rm{1}\ },
        Yingqian Min\textsuperscript{\rm{1}\ },
    } \\
    \textbf{
        Wayne Xin Zhao\textsuperscript{\rm{1}\thanks{\ \ Corresponding author.}},
        Binbin Hu\textsuperscript{\rm{2}\ },
        Ziqi Liu\textsuperscript{\rm{2}\ },
        Zhiqiang Zhang\textsuperscript{\rm{2}\ }
    } \\
    \textsuperscript{1}Gaoling School of Artificial Intelligence, Renmin University of China 
    \textsuperscript{2}Ant Group \\
    \texttt{txy20010310@163.com, wxl1999@foxmail.com}
}

\begin{document}
\maketitle

\begin{abstract}

Recent advancements in long chain-of-thoughts~(long CoTs) have significantly improved the reasoning capabilities of large language models~(LLMs).
Existing work finds that the capability of long CoT reasoning can be efficiently elicited by tuning on only a few examples and can easily transfer to other tasks.
This motivates us to investigate \textit{whether long CoT reasoning is a general capability for LLMs}.
In this work, we conduct an empirical analysis for this question from the perspective of \textit{representation}.
We find that LLMs do encode long CoT reasoning as a general capability, with a clear distinction from vanilla CoTs.
Furthermore, domain-specific representations are also required for the effective transfer of long CoT reasoning.
Inspired by these findings, we propose \OURS, a novel representation engineering method to unleash the \textit{general} long CoT reasoning capabilities of LLMs.
Extensive experiments demonstrate the effectiveness and efficiency of \OURS in both in-domain and cross-domain scenarios.
The code is available at \url{https://github.com/txy77/GLoRE}.

\end{abstract}
\section{Introduction}
\label{sec-introdction}

Recently, slow-thinking reasoning models, such as OpenAI's o1 series of models~\cite{openai-o1} and DeepSeek-R1~\cite{deepseek-r1}, have significantly advanced the capabilities of large language models~(LLMs)~\cite{LLM-survey}.
As a typical approach, these reasoning models leverage long chain-of-thoughts~(long CoTs), encompassing planning, validation, and backtracking strategies, to solve complex reasoning tasks~\cite{Qwen-Model,kimi-rl,Bo-2025-arXiv-BOLT}.
{Most existing work focuses on eliciting long CoTs on tasks that are easy to verify, such as mathematics~\cite{ChainLM,Edward-2025-arXiv-Demystifying} and coding~\cite{Xu-2025-arXiv-RedStar}.}
They find that the capability of long CoT reasoning can be efficiently elicited with only thousands of training examples~\cite{Ye-2025-arXiv-LIMO}.
Furthermore, some recent work finds that this capability can easily transfer to other tasks, even without any task-specific examples~\cite{du2025virgo}.
These interesting phenomena raise a question: \textit{Is long CoT reasoning a \textbf{general} capability encoded in LLMs?}

\begin{figure}[t]
    \centering
    \includegraphics[width=\columnwidth]{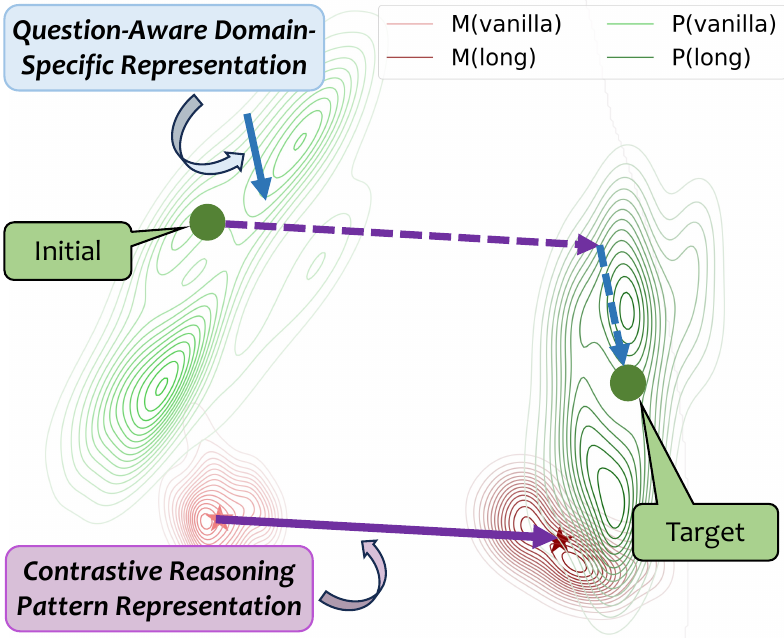}
    \caption{
        The illustration of how \OURS unlocks the general long CoT reasoning capabilities through representation engineering in the parameter space. For a specific problem, we first employ a contrastive reasoning pattern to transition the model from the vanilla CoT area to the long CoT area. Then, we inject domain-specific representations to steer the model toward the precision space tailored for this problem. Here, "M" and "P" denote math and physics, respectively.
    }
\label{fig:intro}
\end{figure}

In this work, we take the first step towards unraveling the mystery from the perspective of \textit{representation engineering}~\cite{Representation-Engineer}.
As a transparent and interpretable method, representation engineering treats representation as the fundamental unit of analysis to understand and control high-level capabilities of LLMs, such as instruction following~\cite{re-instruction}, personality~\cite{re-personality}, and hallucination~\cite{re-hallucination1,re-hallucination2}.
Specifically, representations are extracted from the encodings of LLMs for data that reflect specific capabilities~\cite{Vector-NIPS}.
These representations can then be used for analysis and control of model behaviors.

\begin{figure}[t]
    \centering
    \begin{subfigure}[b]{0.485\columnwidth}
        \centering
        \includegraphics[width=\columnwidth]{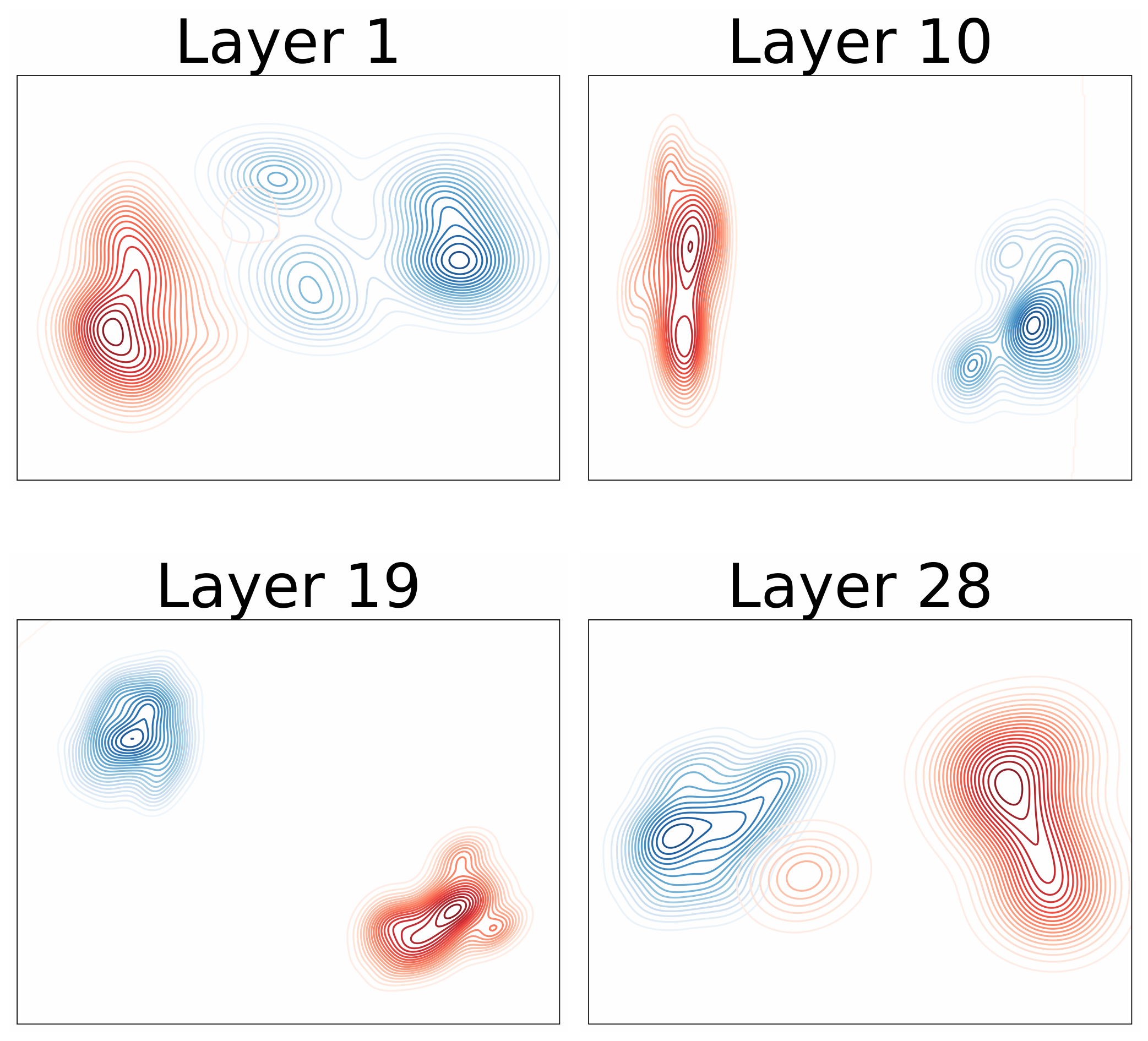}
        \caption{
            Different layers. (blue for vanilla, red for long)
        }
        \label{fig:intro-1}
    \end{subfigure}
    \begin{subfigure}[b]{0.485\columnwidth}
        \centering
        \includegraphics[width=\columnwidth]{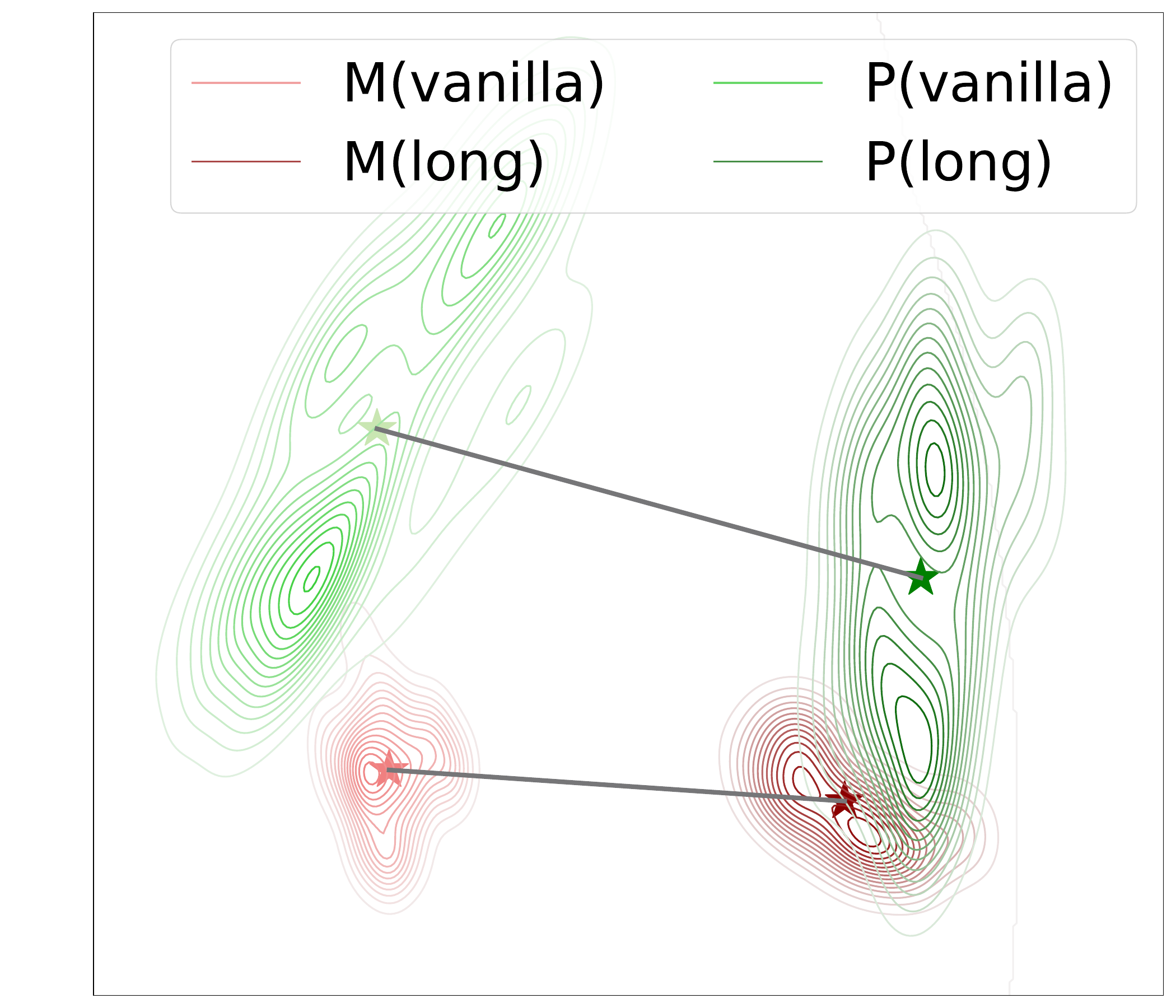}
        \caption{
            Different domains. (``M'' for math, ``P'' for physics)
        }
        \label{fig:intro-2}
    \end{subfigure}
    \caption{
        Visualization of vanilla and long CoTs on Qwen2.5-7B-Instruct.
    }
\label{fig:empirical}
\end{figure}

Inspired by this approach, we leverage representation engineering to analyze the mechanism of long CoT reasoning.
As illustrated in Figure~\ref{fig:intro-1}, the representations of long CoTs across diverse problems are concentrated in a specific area of the whole space.
In addition, their distribution areas are clearly distinct from those of vanilla CoTs.
Taken together, the two pieces of evidence suggest that LLMs do encode long CoT reasoning as a separate and general capability within their parameter spaces.
Based on this insight, we further examine the representations of long and vanilla CoTs across various domains.
The results in Figure~\ref{fig:intro-2} show that different domains share similar contrastive representations between long and vanilla CoTs, which further demonstrates the transferability of long CoT reasoning.
In addition, the representations of mathematical domains are relatively concentrated, while those of other domains (\eg physics) are more dispersed.
This suggests that general long CoT reasoning requires not only \textit{unique reasoning patterns} but also \textit{domain-specific information}.
That is, domain-specific long CoT data is important for the elicitation of long CoT reasoning in specific domains.
However, not all domains are easy to construct high-quality long CoTs.

To facilitate \textbf{G}eneral \textbf{Lo}ng CoT reasoning across domains, we further propose a \textit{training-free} approach based on \textbf{R}epresentation \textbf{E}ngineering, namely \OURS.
Specifically, we first construct the representations of long CoT patterns from contrastive representations between long and vanilla CoT data of high-resource domains (\ie mathematics).
Then, we build a domain-specific representation memory by using vanilla CoT data from corresponding domains.
At inference time, we first retrieve relevant domain-specific representations from the corresponding memory and then inject both the retrieved representations and those of long CoT patterns into the LLM for reasoning.
Such an approach is \textit{cost-efficient}, as it is free from training and only relies on long CoT data from high-resource domains.
To validate the effectiveness of our approach, we conduct experiments in both in-domain~(mathematics) and cross-domain scenarios~(GPQA, including physics, chemistry, and biology).
In particular, our approach consistently outperforms all the training-free baselines and even surpasses the supervised fine-tuning method, while maintaining lower time complexity.

Our contributions can be summarized as follows:

$\bullet$ To the best of our knowledge, we are the first to analyze the mechanism of long CoT reasoning from the perspective of representation.

$\bullet$ We propose a novel training-free method based on representation engineering, which can effectively unlock the general long CoT reasoning capabilities of LLMs.

$\bullet$ Extensive experiments demonstrate the effectiveness and efficiency of our proposed method in both in-domain and cross-domain scenarios.

\section{Related Work}
\label{sec-related_work}

Our work is related to the following two research directions.

\paratitle{Large Language Model Reasoning.}
Recently, improving the reasoning capabilities of LLMs has become a critical challenge.
Prior approaches, such as test-time search~\cite{rest-search,DAWN-ICL,rstar-search,HaluSystem}, distillation~\cite{Distill1,STILL2}, and reinforcement learning~\cite{deepseek-r1}, enable LLMs to engage in deliberate thinking~\cite{GPO,ChainLM,RL-RAG}.
Despite their remarkable success, the underlying mechanisms of LLM reasoning remain unclear.
Some studies~\cite{MathNeuro,ACL-Neurocot} analyze this by localizing specific neurons, but they only focus on isolated neuron connections, neglecting the cooperative activity of multiple neurons.
Other work~\cite{RoT,ICLR-recot} addresses this via representation engineering to better control neuron collaboration.
However, these studies are often limited to short-form CoT, struggling to fully unlock the reasoning potential of LLMs.
In this paper, we focus on exploring the mechanism of long CoT reasoning through representation engineering.

\paratitle{Representation Engineering.}
Representation engineering~\cite{Representation-Engineer} treats internal representations as the fundamental unit, focusing on analyzing and manipulating them within neural networks.
As a well-established technique, it has been applied in various areas such as personality modeling~\cite{re-personality}, instruction following~\cite{re-instruction}, hallucination alleviation~\cite{re-hallucination1,re-hallucination2,HADES}, and safety improvement~\cite{in-context-vector-icml24}.
While prior work focuses on simple concepts like sentiment~\cite{re-sentiment} and style~\cite{re-style1,re-style2}, our work aims to address the more complex challenge: understanding and unlocking general long CoT reasoning capabilities of LLMs.

\section{Empirical Analysis}
\label{sec:preliminary}

In this section, we first introduce the background of representation engineering and then use it to conduct an empirical analysis of long CoT reasoning.

\subsection{Background: Representation Engineering}
\label{sec:re}

The Hopfieldian view~\cite{hopfield-neural} explains cognition and behavior as emerging from transformations or movements within neural populations in response to external stimuli.
Building upon this perspective, representation engineering~\cite{Representation-Engineer} is proposed, which is a widely used approach for the mechanism interpretability of LLMs.
It treats representations as the fundamental unit of various mechanisms in LLMs for analysis.
This approach primarily encompasses two components: representation extraction and control.
We will detail them in the following part.

\paratitle{Representation Extraction.}
It focuses on identifying high-level concepts or functions encoded in LLMs.
For a typical Transformer~\cite{Transformers-Architecture} model, the outputs of multi-head attention~(MHA), multi-layer perception~(MLP), and hidden states can all be considered as representations, with each connected through the residual stream.
At a given layer $l$ and token position $t$, the hidden state $h^t_l$ is computed recursively as follows:
\begin{equation}
\label{eq:representation}
h_l^t = h_{l-1}^t + a^t_l + m^t_l,
\end{equation}
where $a^t_l$ and $m^t_l$ represent the outputs from MHA and MLP, respectively.
Here, we follow \citet{Representation-Engineer} to extract representations from the hidden states at the final token position due to the sequential nature of language modeling.

\paratitle{Representation Control.}
It aims to steer model behaviors with extracted representations.
This process typically first establishes a representation controller to modulate extracted representations.
Then, the controller will inject the representations of target behaviors into the representations of LLMs.
Here, we follow \citet{task-vector} to utilize a linear module as the representation controller and select a specific layer for representation injection.
Such a method can achieve fine-grained control of model behaviors while preserving efficiency.

\subsection{Analysis of Long CoT Representations}
\label{subsec:analysis-thinking-system}

In this part, we first describe how to extract long CoT representations and then conduct an empirical analysis about them.

\paratitle{Extraction of Representations.}
To extract representations, first, we prompt an LLM to collect its vanilla CoTs $s_i$ and long CoTs $l_i$ for a set of questions $x_i \in \mathcal{X}$.
Then, we concatenate each problem with the corresponding CoT and input this into the LLM for encoding.
As stated in Section~\ref{sec:re}, the hidden states of the layer $L$ at the final token position are extracted as the representations, which can be represented as follows:
\begin{equation}
R_L(s_i) = h_L^{-1}(x_i;s_i) \ \
R_L(l_i) = h_L^{-1}(x_i;l_i),
\end{equation}
where $h_L^{-1}(s)$ denotes the hidden states of the string $s$ at the last token position and layer $L$, and $;$ denotes string concatenation.
After performing the above operation, we can obtain a set of representations for vanilla and long CoTs.

\paratitle{Analysis of General Representations.}
To analyze the characteristics of vanilla and long CoTs, we visualize their representations to compare their distributions.
Specifically, we employ a dimensionality reduction approach (\ie t-SNE~\cite{t-sne}) to map representations obtained from the above part onto a 2D plane.
As illustrated in Figure~\ref{fig:intro-1} (more figures in Appendix~\ref{app:all-representations}), the representations of various long CoTs are concentrated in a specific area of the whole space.
In addition, their distribution areas are clearly distinct from those of vanilla CoTs.
Taken together, the two pieces of evidence suggest that LLMs do encode long CoT reasoning as a separate general capability in their parameter spaces.
Moreover, we find that the separation between these two types of CoTs is the most pronounced in the middle layers of the model, while less clear in the early and final layers.
This phenomenon may be attributed to the fact that middle layers integrate information from early layers and are more informative~\cite{Skean-2025-Layer-arXiv}, playing a critical role in capturing high-level concepts (\eg CoT reasoning) (See Appendix~\ref{app:qa-CoT}).

\paratitle{Analysis of Domain-Specific Representations.}
In this part, we further examine the characteristics of vanilla and long CoTs in specific domains.
Specifically, we collect representations in mathematical and other domains (\ie physics, chemistry, and biology) and visualize them following the previous part.
As shown in Figure~\ref{fig:intro-2} (more figures in Appendix~\ref{app:all-representations-domain}), different domains share similar contrastive representations between long and vanilla CoTs, which further demonstrates the transferability of long CoT reasoning.
In addition, the representations of mathematical domains are relatively concentrated, while those of other domains (\eg physics) are more dispersed.
This may be due to the fact that mathematical problems focus on logical reasoning patterns, while problems in other domains also require domain-specific information.
That is, domain-specific CoT data plays an important role for the elicitation of long CoT reasoning within these domains~\cite{EasyEP}.

\section{Unlocking General Long CoT Reasoning Capabilities}
\label{sec:method}

\begin{figure*}[t]
    \centering
    \includegraphics[width=\textwidth]{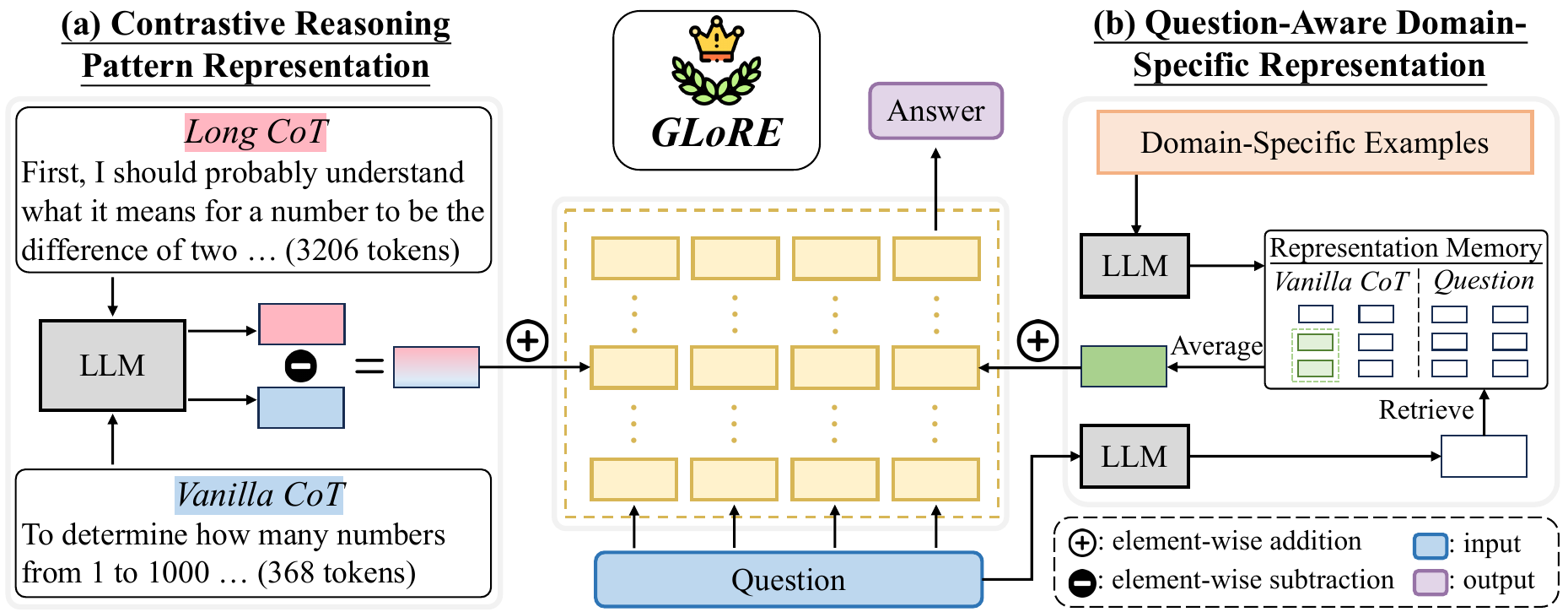}
    \caption{
        The overview of our approach \OURS.
        It extracts contrastive reasoning pattern representations (part a) and question-aware domain-specific representations (part b) and injects them into LLMs.
        For part a, we derive the representations from the difference between long and vanilla CoTs.
        For part b, we construct a domain-specific representation memory from vanilla CoTs only and retrieve representations relevant to the question.
    }
\label{fig:main}
\end{figure*}

As discussed in Section~\ref{subsec:analysis-thinking-system}, long CoT reasoning appears to be a general capability potentially encoded in LLMs.
Therefore, it is feasible to unlock this capability through representation engineering with long-CoT data.
However, not all domains are easy to obtain high-quality long CoT data.
To solve this, our idea is to decouple long CoT reasoning into general reasoning patterns and domain-specific information.
Since both reasoning patterns and domain-specific information are important for general long CoT reasoning, we design tailored methods to extract each kind of representation and inject them to control model behaviors.
The overall framework is illustrated in Figure~\ref{fig:main}.

\subsection{Contrastive Reasoning Pattern Representation}
\label{sec:general-pattern}

Recall that different domains share similar contrastive representations between long and vanilla CoTs, as stated in Section~\ref{subsec:analysis-thinking-system} (``Analysis of General Representations'').
Motivated by this finding, we propose to facilitate the transition from vanilla CoT to long CoT patterns by leveraging contrastive representations from high-resource domains (\eg mathematics).
In the following part, we introduce how to extract and inject contrastive reasoning pattern representations.

\paratitle{Representation Extraction.}
Given a set of questions $\mathcal{X}$ from high-resource domains, first, we extract their representations following the method in Section~\ref{subsec:analysis-thinking-system}.
Then, to enhance the transferability of extracted representations, we average the contrastive representations across all the data, abstracting away domain-specific information.
Formally, we calculate the contrastive reasoning pattern representation $p_L$ at the layer $L$ as follows:
\begin{equation}
p_L = \frac{1}{|\mathcal{X}|}\sum_{i=1}^{|\mathcal{X}|}(R_L(l_i)-R_L(s_i)),
\end{equation}
where $\mathcal{X}$ is the set of questions, $R_L(l_i)$ and $R_L(s_i)$ denote the representations of long and vanilla CoT at layer $L$, respectively.

\paratitle{Representation Control.}
After extracting the reasoning pattern representation, we inject it into specific layers of LLMs during the forward pass to guide LLMs towards deliberate reasoning.
Since the reasoning pattern is a global behavior in the generation process, we choose to inject the representation into that of the first token, ensuring that each following token can attend to it.
In addition, to preserve the original capabilities of LLMs as much as possible, we follow \citet{in-context-vector-icml24} to normalize the updated representations.
Formally, the updated representation $\tilde{h}_L^0$ of the first token at the layer $L$ is calculated as follows:
\begin{align}
\tilde{h}_L^0 &= h_L^0 + \lambda_p \cdot p_L , \\
\tilde{h}_L^0 &= \tilde{h}_L^0 \cdot \frac{\Vert h_L^0 \Vert_2}{\Vert \tilde{h}_L^0 \Vert_2},
\end{align}
where $\lambda_p$ is the hyperparameter controlling the strength of injection.

\subsection{Question-Aware Domain-Specific Representation}
\label{subsec:domain-representation}

After injecting the representation of the long CoT reasoning pattern, LLMs can be steered to generate long CoTs at inference.
However, as stated in Section~\ref{subsec:analysis-thinking-system} (``Analysis of Domain-Specific Representations''), reasoning patterns only are not enough.
For effective long CoT reasoning, domain-specific information is also important.
Therefore, we propose to extract domain-specific representations and construct a representation memory.
Note that the memory is constructed from vanilla CoT data, which is easy to obtain.
At inference time, to provide domain-specific information for long CoT reasoning, we retrieve representations relevant to the question and inject them into LLMs.
In the following part, we detail these two components.

\paratitle{Domain-Specific Representation Memory.}
Since domain-specific information is shared across long and vanilla CoTs, we propose to collect them using only vanilla CoTs, which can be easily obtained through methods like zero-shot prompting (\eg ''Let's think step by step.``).
Then, we can construct a representation memory for relevant information retrieval at inference time.
Specifically, for a question $x_i$ and its associated vanilla CoT $s_i$, we extract the representation of the question $R_L(x_i)$ as the key of the memory and the representation of the question combined with the CoT $R_L(s_i)$ as the value of the memory.

\paratitle{Question-Aware Representation Retrieval.}
With the domain-specific representation memory, we can retrieve representations relevant to the specific question for better long CoT reasoning.
Specifically, for a question $x$, we extract its representation as the query to retrieve top-$k$ representations from the memory.
The retrieval is implemented by first calculating cosine similarity between the query and keys and then extracting the corresponding values with the highest similarity values.
To highlight common information, we further average the $k$ retrieved representations.
Finally, we follow the method in Section~\ref{sec:general-pattern} to inject the domain-specific representation $d$ into LLMs.
Different from Section~\ref{sec:general-pattern}, we choose to inject into the final token position, as it can influence the generation of the next token while preserving the encoding of previous tokens.
It can be represented as follows:
\begin{align}
\tilde{h}_L^{-1} &= h_L^{-1} + \lambda_d \cdot d , \\
\tilde{h}_L^{-1} &= \tilde{h}_L^{-1} \cdot \frac{\Vert {h}_L^{-1} \Vert_2}{\Vert \tilde{h}_L^{-1} \Vert_2},
\end{align}
where $\lambda_d$ is the hyperparameter controlling the strength of injection.
\section{Experiments}
\label{sec-experiment}

In this section, we first set up the experiments, then report the results and conduct a detailed analysis.

\begin{table*}[t]
\centering
\resizebox{\textwidth}{!}{
\begin{tabular}{cc|cccc|cccc}
    \toprule
    \multicolumn{2}{c|}{{\textbf{Scenarios}}} & \multicolumn{4}{c|}{{\textbf{In-domain}}} & \multicolumn{4}{c}{{\textbf{Cross-domain}}} \\
    \midrule
    \multicolumn{2}{c|}{\multirow{2.5}{*}{\textbf{Task}}} & \multicolumn{4}{c|}{\textbf{Math BenchMarks}} & \multicolumn{4}{c}{\textbf{GPQA}} \\
    \cmidrule{3-10}
    & & \textbf{MATHOAI} & \textbf{AIME24} & \textbf{AMC23} & \textbf{Average} & \textbf{Physics} & \textbf{Chemistry} & \textbf{Biology} & \textbf{Overall} \\ 
    \midrule
    \multirow{8}{*}{\begin{tabular}[c]{@{}c@{}}\textbf{Qwen2.5-7B}\\\textbf{-Instruct}\end{tabular}}      
    & Zero-shot CoT                                     & 72.80 & 13.33 & 47.50 & 44.54 & 38.37 & 22.58 & 36.84 & 30.81 \\ 
    & Few-shot CoT                                      & 69.80 & 6.67  & 42.50 & 39.66 & 33.72 & 20.43 & 26.32 & 26.77 \\
    & BoostStep                                         & 70.80 & 10.00 & 45.00 & 41.93 & 34.88 & 23.66 & 21.05 & 28.28 \\
    & MathNeuro                                         & 73.60 & 16.67 & 50.00 & 46.76 & 37.21 & 22.58 & 36.84 & 30.30 \\
    & RoT                                               & 73.80 & 16.67 & 52.50 & 47.66 & 41.86 & 23.66 & 42.11 & 33.33 \\
    & SFT                                               & 74.80 & 23.33 & \textbf{60.00} & 52.71 & 44.19 & 23.66 & \textbf{52.63} & 35.35 \\
    \cmidrule{2-10}
    & \OURS                                             & \textbf{76.20} & \textbf{26.67} & \textbf{60.00} & \textbf{54.29} & \textbf{46.51} & \textbf{25.81} & 42.11 & \textbf{36.36} \\
    \midrule
    \multirow{8}{*}{\begin{tabular}[c]{@{}c@{}}\textbf{Llama3.1-8B}\\\textbf{-Instruct}\end{tabular}}      
    & Zero-shot CoT                                     & 48.20 & 3.33 & 30.00 & 27.18 & 19.77 & 16.13 & 42.11 & 20.20 \\ 
    & Few-shot CoT                                      & 45.60 & 6.67 & 27.50 & 26.59 & 22.09 & 15.05 & 36.84 & 20.20 \\
    & BoostStep                                         & 47.60 & 6.67 & 25.00 & 26.42 & 20.93 & 18.28 & 36.84 & 21.21 \\
    & MathNeuro                                         & 49.00 & 10.00 & 30.00 & 29.67 & 22.09 & 22.58 & \textbf{47.37} & 24.75 \\
    & RoT                                               & 49.40 & 10.00 & 32.50 & 30.63 & 24.42 & 23.66 & 42.11 & 25.76 \\
    & SFT                                               & 50.20 & 13.33 & \textbf{35.00} & 32.84 & 26.88 & 27.96 & \textbf{47.37} & 29.29 \\
    \cmidrule{2-10}
    & \OURS                                             & \textbf{51.60} & \textbf{16.67} & \textbf{35.00} & \textbf{34.42} & \textbf{27.91} & \textbf{29.03} & \textbf{47.37} & \textbf{30.30} \\  
    \bottomrule
    \end{tabular}
    
}
\caption{Performance comparison in both in-domain and cross-domain scenarios using Qwen2.5-7B-Instruct and Llama3.1-8B-Instruct. The best method in each group is marked in \textbf{bold}.}
\label{tab:main}
\end{table*}

\subsection{Experimental Setup}
\label{subsec:experiment_setup}

\paratitle{CoT Examples Construction.}
To obtain vanilla and long CoT examples, we utilize open-source data from STILL-2~\cite{STILL2}, which is a high-quality dataset distilled from DeepSeek-R1-Lite-Preview~\cite{deepseek-r1}.
From this dataset, we randomly select 100 examples from the mathematics, physics, chemistry, and biology domains, respectively.

\paratitle{Datasets.}
To comprehensively evaluate the efficacy of our proposed method, we conduct experiments in two scenarios: in-domain and cross-domain.
For the in-domain scenario, we evaluate our method on several challenging open-source mathematical benchmarks, including MATHOAI~\cite{MATH-Dataset}, AIME2024, and AMC2023.
For the cross-domain scenario, we utilize the GPQA~\cite{GPQA-Dataset} dataset, which is a challenging multiple-choice benchmark crafted by domain experts in physics, chemistry, and biology.
In this paper, we use the highest quality diamond set for evaluation following \citet{search-o1}.

\paratitle{Baselines.}
To facilitate a systematic comparison, we select several representative methods, including prompting-based approaches~(\ie Zero-shot CoT, Few-shot CoT, and BoostStep~\cite{arXiv-Zhang-BoostStep}), neuron activation method~(\ie MathNeuro~\cite{MathNeuro}), representation engineering method~(\ie RoT~\cite{RoT}), and supervised fine-tuning method.
Detailed descriptions of these baselines are provided in Appendix~\ref{app:baselines}.

\paratitle{Implementation Details.}
In our experiments, we use two representative open-source LLMs: Qwen2.5-7B-Instruct~\cite{Qwen-Model} and Llama3.1-8B-Instruct~\cite{llama3.1-8b-model}.
For the contrastive reasoning pattern representation, we set the injection strength $\lambda_p$ as 0.1.
For the question-aware domain-specific representation, we set the number of retrieved representations $k$ as 8 and the injection strength $\lambda_d$ as 0.1.
Both representations are injected into the intermediate layer of LLMs.
Following the existing work~\cite{POPE,REAR,HaluAgent}, we use the greedy decoding strategy for inference.

\subsection{Experimental Results}

\begin{figure}[t]
    \centering
    \begin{subfigure}[b]{0.49\columnwidth}
        \centering
        \includegraphics[width=\columnwidth]{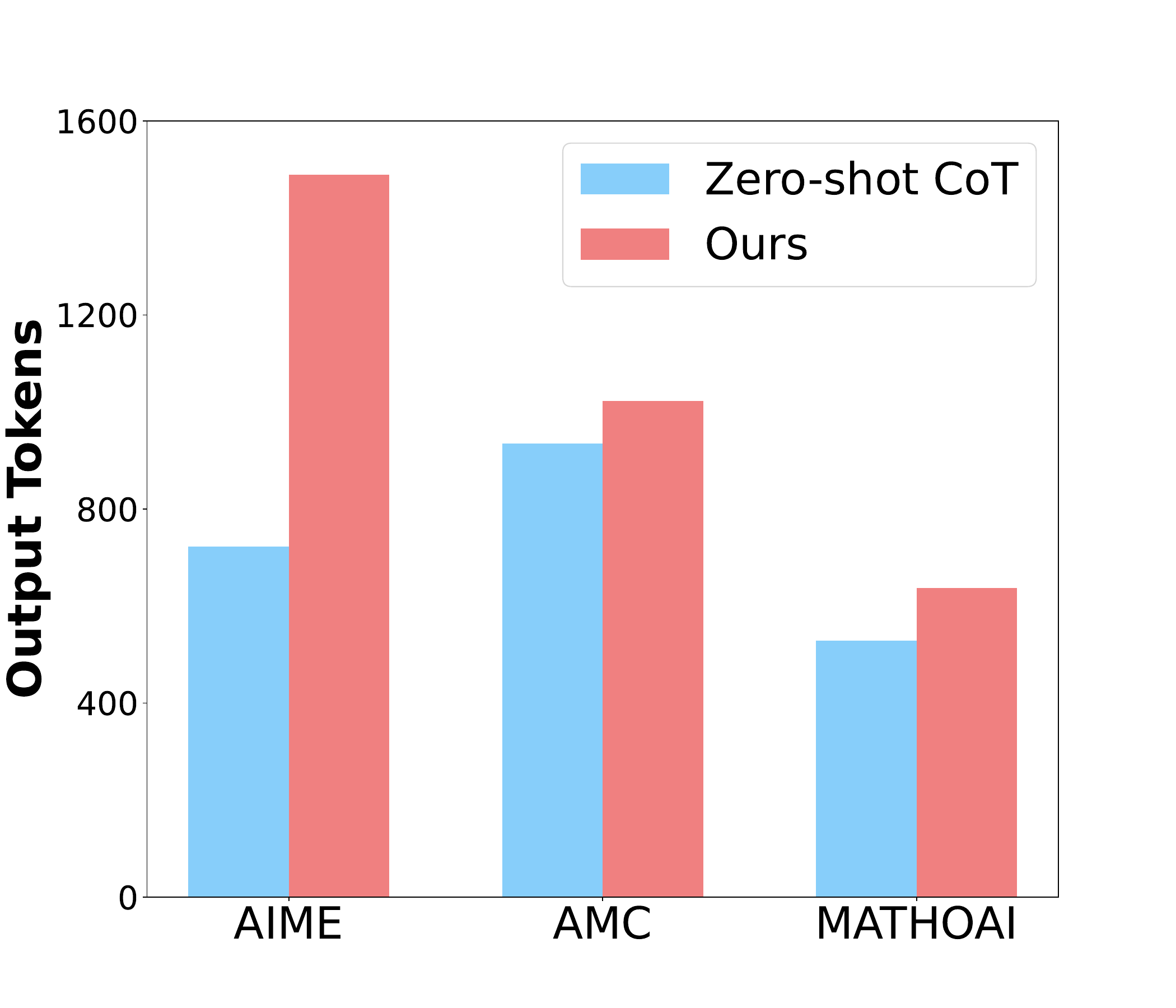}
        \caption{In-domain}
        \label{fig:in-domain-length}
    \end{subfigure}
    \begin{subfigure}[b]{0.49\columnwidth}
        \centering
        \includegraphics[width=\columnwidth]{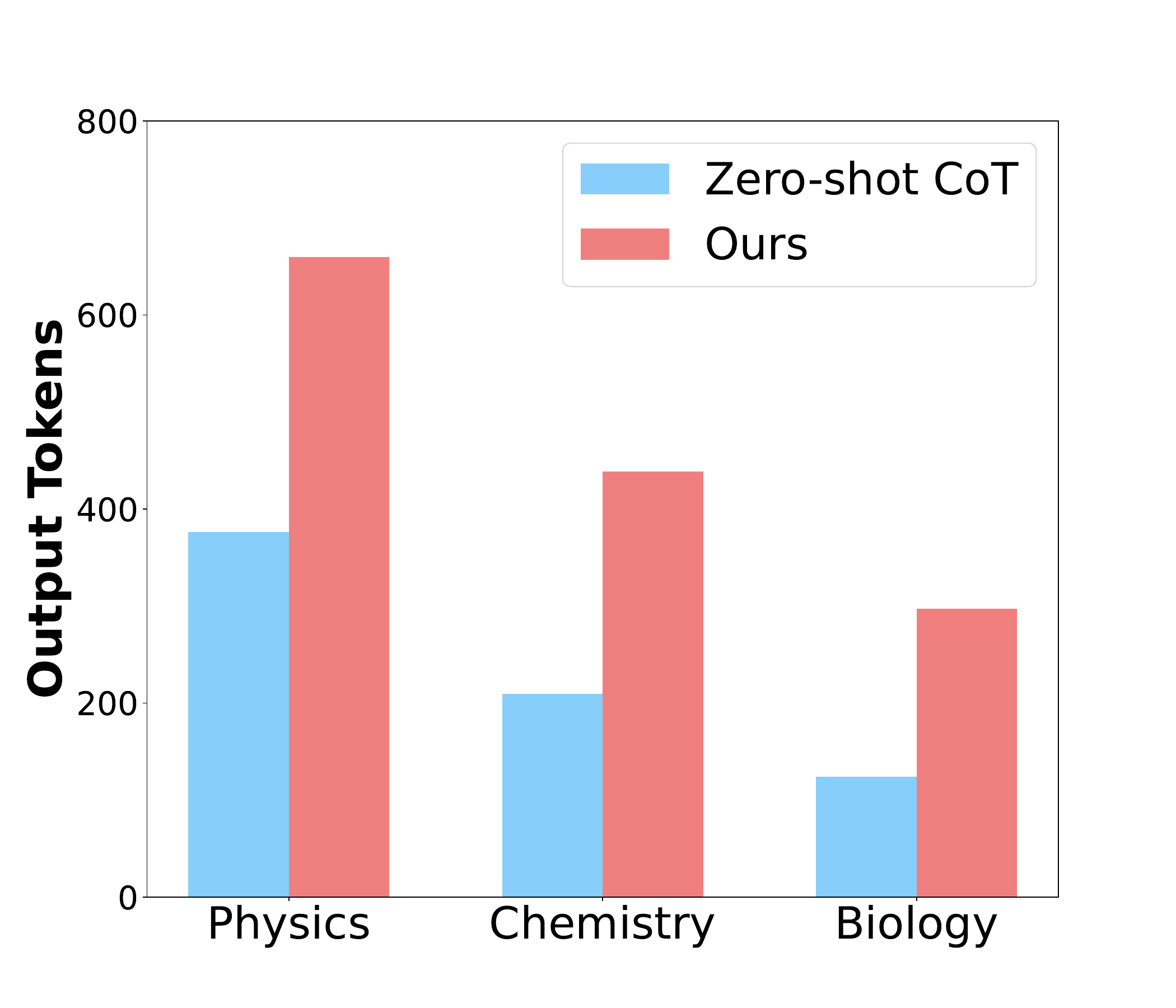}
        \caption{Cross-domain}
        \label{fig:cross-domain-length}
    \end{subfigure}
    \caption{The average number of output tokens generated by Qwen2.5-7B-Instruct.}
\label{fig:output_tokens}
\end{figure}

The experimental results are presented in Table~\ref{tab:main}.
As we can see, the few-shot CoT method~(\ie few-shot CoT and BoostStep) performs poorly, even worse than the zero-shot CoT.
The main reason is that LLMs lose their ability to learn from demonstrations after supervised fine-tuning~\cite{wei2023instructiongpt-4}.
Adding examples at the discrete token level can increase input length, which distracts the model from the current problem and disrupts the inference process~\cite{BAMBOO,CAFE}.
In contrast, MathNeuro can improve performance by identifying and scaling relevant neurons, but it focuses solely on specific neuron connections, neglecting the cooperative activity of multiple neurons.
To address this limitation, RoT introduces contrastive representations of CoT and non-CoT prompts, enabling fine-grained control over the reasoning process.
However, this approach is insufficient to guide the model toward deliberate thinking, resulting in limited performance improvements.

Finally, \OURS significantly outperforms all the training-free baselines and even surpasses the supervised fine-tuning method.
{Our approach first uses a contrastive reasoning pattern representation to switch the LLM to a slow thinking pattern, enabling it to engage in deep thinking and perform step-by-step reasoning.}
This allows \OURS to generate longer and more detailed reasoning solutions, indicating that the representation effectively guides LLMs into a slow-thinking mode, as illustrated in Figure~\ref{fig:output_tokens}.
Additionally, for specific problems, we leverage question-aware domain-specific representation, which provides domain-specific information during inference to achieve fine-grained control over the reasoning process.

\subsection{Detailed Analysis}

In this part, we construct a detailed analysis of the effectiveness and efficiency of our approach.

\subsubsection{Ablation Study}

\begin{table}[t]
\centering
\resizebox{\linewidth}{!}{
\begin{tabular}{l|cc|cc}
\toprule
\textbf{Task} & \multicolumn{2}{c|}{\textbf{MATHOAI}} & \multicolumn{2}{c}{\textbf{GPQA}} \\
\midrule
\textbf{Model} &  \begin{tabular}[c]{@{}c@{}}\textbf{Qwen2.5-7B}\\ \textbf{-Instruct}\end{tabular} &
\begin{tabular}[c]{@{}c@{}}\textbf{Llama3.1-8B}\\ \textbf{-Instruct}\end{tabular} & 
\begin{tabular}[c]{@{}c@{}}\textbf{Qwen2.5-7B}\\ \textbf{-Instruct}\end{tabular} & \begin{tabular}[c]{@{}c@{}}\textbf{Llama3.1-8B}\\ \textbf{-Instruct}\end{tabular} \\
\midrule
\OURS       & 76.20 & 51.60 & 36.36 & 30.30 \\
\midrule
w/o CR     & 73.40 & 49.80 & 33.84 & 25.76 \\
w VR        & 73.60 & 50.00 & 33.84 & 26.26 \\
w LR        & 75.40 & 51.20 & 35.35 & 30.30 \\
\midrule
w/o DR      & 74.20 & 50.60 & 32.32 & 26.77 \\
w LDR       & 76.40 & 51.60 & 37.88 & 29.80 \\
\bottomrule
\end{tabular}
}
\caption{Ablation study on MATHOAI and GPQA datasets. ``CR'', ``VR'', ``LR'', ``DR'' and ``LDR'' denote contrastive reasoning pattern representation, vanilla CoT representation, long CoT representation, question-aware domain-specific vanilla, and long CoT representation.}
\label{tab:ablation}
\end{table}

Our approach incorporates two key components to activate the long CoT reasoning capabilities of LLMs.
To validate each component of our proposed method, we conduct an ablation study by removing or replacing the contrastive reasoning pattern and question-aware domain-specific representation on MATHOAI and GPQA datasets.

The results are presented in Table~\ref{tab:ablation}.
We can see that removing any component would lead to performance degradation, indicating that all the components in our method are helpful.
Specifically, {for the contrastive reasoning pattern representation}, we compare the effects of injecting only the representation of vanilla CoT or long CoT and observe that both lead to performance degradation.
In particular, injecting only vanilla CoT representation significantly reduces performance, as the model fails to transition into a slow-thinking mode.
For the {question-aware domain-specific representation}, we observe that injecting long CoT domain-specific thought achieves performance comparable to vanilla CoT.
This indicates that our method can effectively leverage vanilla CoT from other domains, highlighting its cost efficiency.

\subsubsection{The Efficiency of \OURS}

\begin{table}[t]
\centering
\resizebox{\linewidth}{!}{
\begin{tabular}{l|cccc}
\toprule
Methods & Zero-shot & Few-shot & BoostStep & \textbf{\OURS} \\
\midrule
T.C. & $\mathcal{O}(p^2)$ & $\mathcal{O}((d+p)^2)$ & $\mathcal{O}(n(d+p)^2)$ & $\mathcal{O}(p^2)$ \\
\bottomrule
\end{tabular}
}
\caption{The efficiency analysis of \OURS and previous work. Here, ``T.C.'' is the time complexity, $d$, $p$ and $n$ denote the length of the demonstrations, the length of the problem and the number of the reasoning steps.}
\label{tab:efficiency}
\end{table}

In this part, We discuss the efficiency of \OURS, as shown in Table~\ref{tab:efficiency}.
First, the few-shot CoT method incorporates additional demonstrations in the input, which leads to the increased inference time.
This is because the time complexity of the Transformer scales quadratically with the length of the input sequence~\cite{LongReD,ZYL,DenoiseVAE}.
Additionally, BoostStep decomposes the reasoning process into multiple substeps and guides the model with relevant examples at each step.
This requires the model to perform multiple reasoning iterations for a single problem, further increasing the computational overhead.
In contrast, \OURS maintains the same time complexity as zero-shot CoT, which is significantly lower than few-shot CoT and BoostStep.
We extract representation from the model's latent space and inject them during inference, without reducing the reasoning efficiency.
This demonstrates that \OURS can significantly enhance reasoning capabilities through representation engineering while preserving computational efficiency.

\begin{figure*}[t]
    \centering
    \includegraphics[width=0.9 \textwidth]{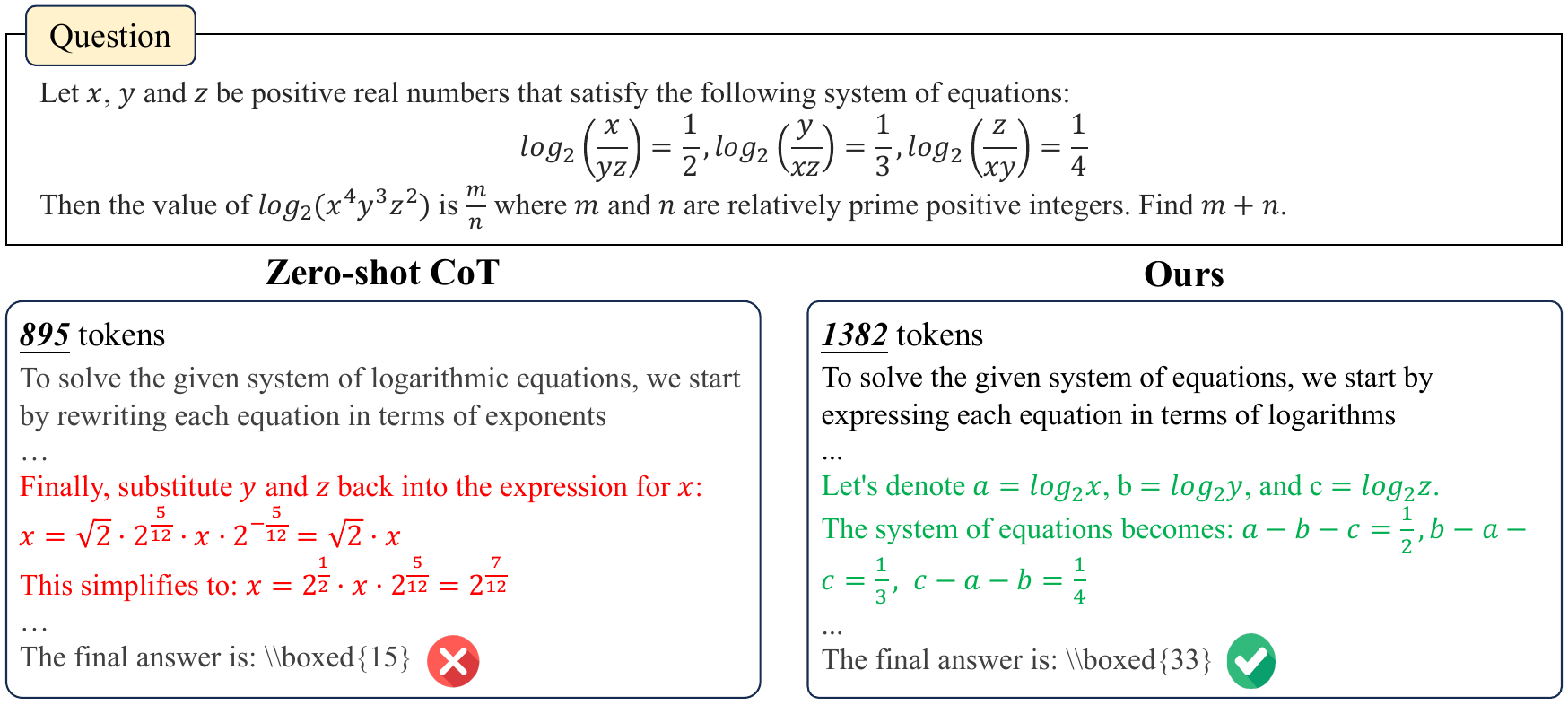}
    \caption{A specific example of how our method activates the long CoT reasoning capabilities of LLMs.}
\label{fig:case}
\end{figure*}

\subsubsection{Hyper-parameters Analysis}

\begin{figure}[t]
    \centering
    \begin{subfigure}[b]{0.49\columnwidth}
        \centering
        \includegraphics[width=\columnwidth]{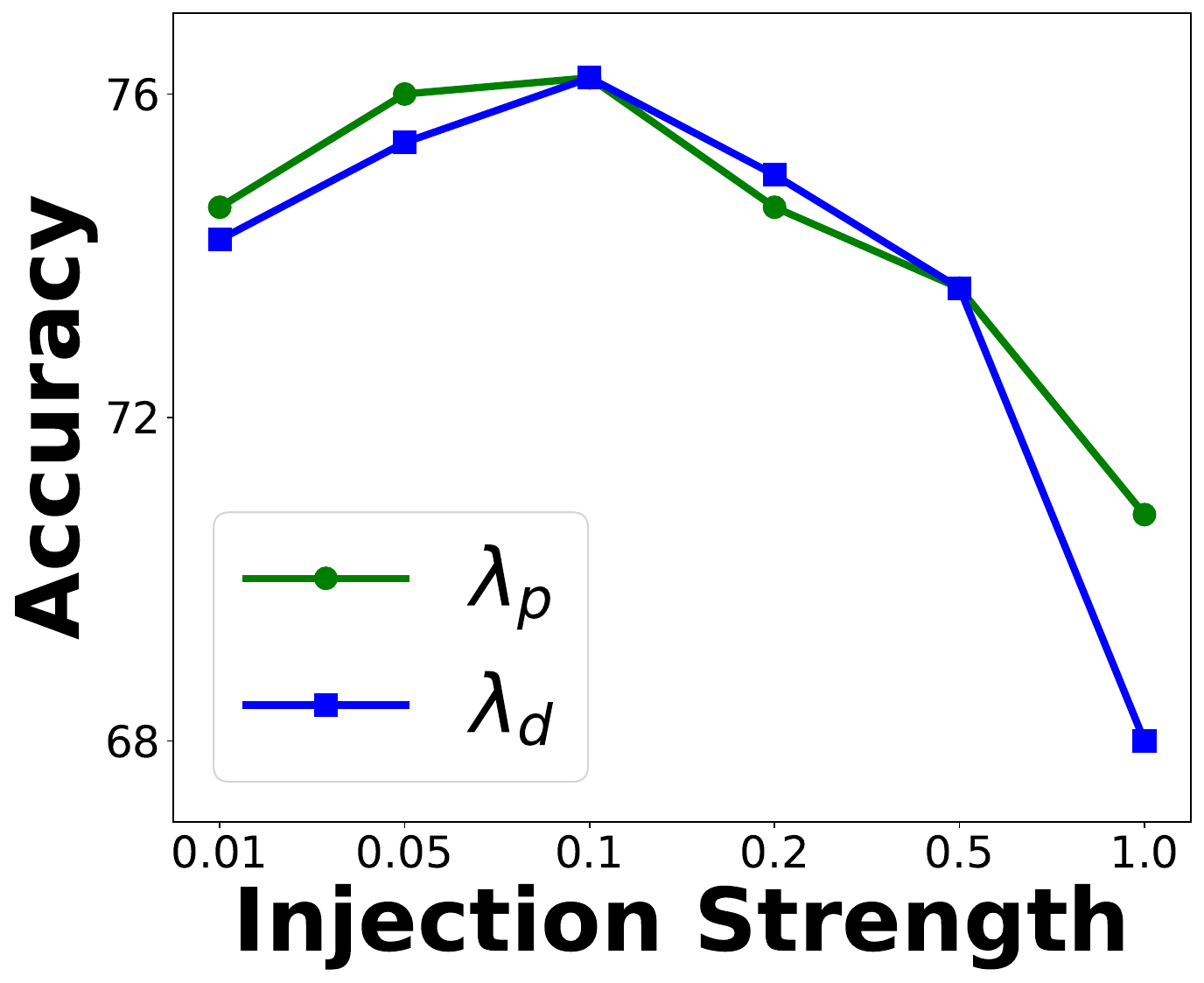}
        \caption{MATHOAI}
        \label{fig:inject_strength}
    \end{subfigure}
    \begin{subfigure}[b]{0.49\columnwidth}
        \centering
        \includegraphics[width=\columnwidth]{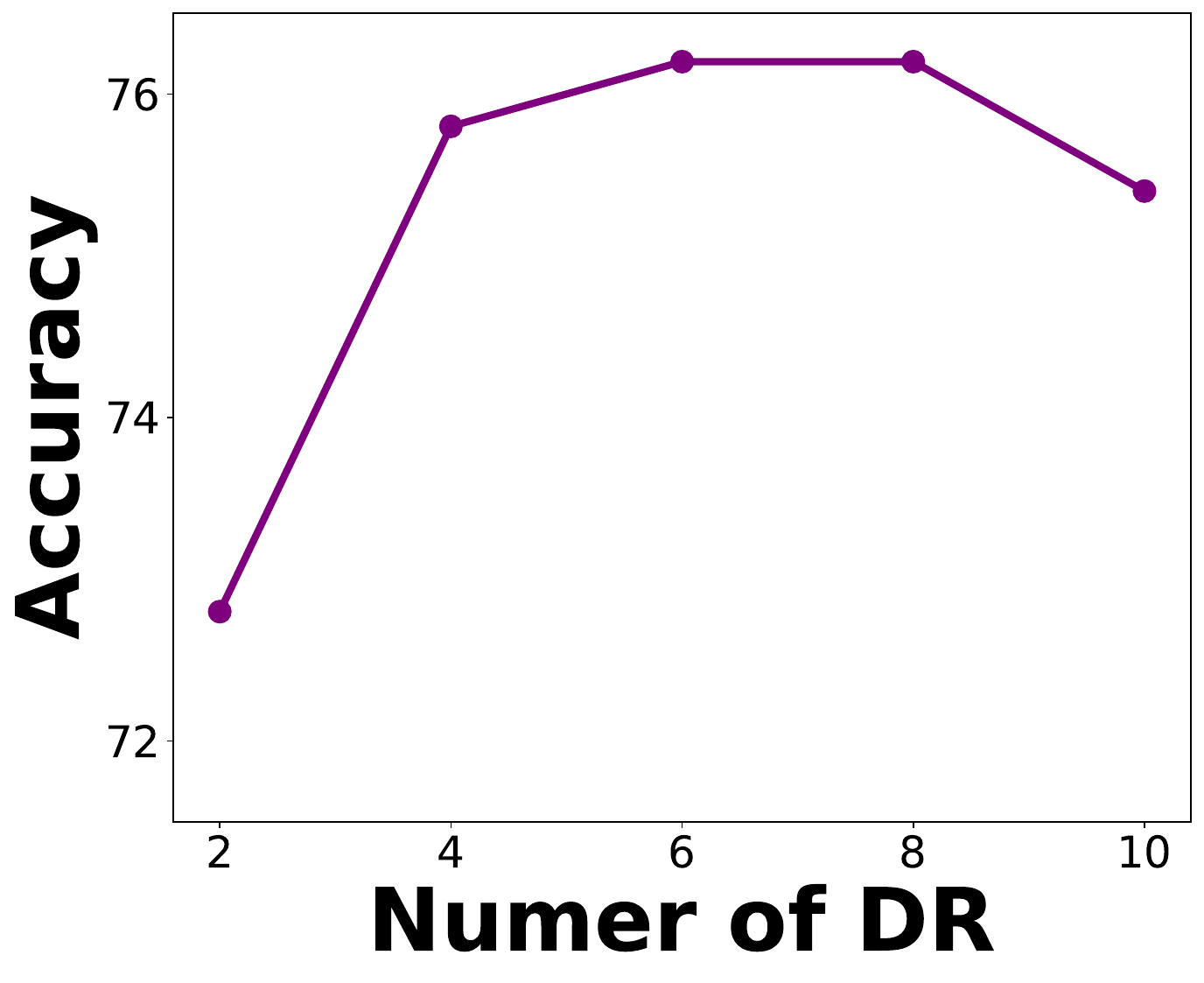}
        \caption{GPQA}
        \label{fig:k_impact}
    \end{subfigure}
    \caption{Performance comparison \emph{w.r.t.} the inject strength $\lambda_p$ and $\lambda_d$, and the number of retrieved domain-specific representations $k$ on the MATHOAI dataset using Qwen2.5-7B-Instruct. Here, ``DR'' denotes the retrieved question-aware domain-specific representation.}
\label{fig:hyper-parameter}
\end{figure}

\OURS includes a few hyper-parameters to tune.
In this part, we report the tuning results of three hyper-parameters: the injection strength for contrastive reasoning pattern representation ($\lambda_p$) and question-aware domain-specific representation ($\lambda_d$), and the number of retrieved representations $k$.
The results are shown in Figure~\ref{fig:hyper-parameter}.

We find that the performance is optimal when both injection strengths are set to 0.1.  
If the injection strength is too small, the model cannot effectively perceive the intervention of the representations, preventing it from engaging in slow thinking or incorporating domain-specific information.
Conversely, if the injection strength is too large, the injected representations may disrupt the original semantic information of the model, leading to performance degradation.
Additionally, we observe that \OURS achieves the best performance when the number of similar representations is set to 8.  
If the number is too small, the model cannot access sufficient domain-specific information to support reasoning.
In contrast, if the number is too large, irrelevant information may be introduced, which can interfere with the reasoning process.

\subsubsection{Experiments on Larger Models}

\begin{figure}[t]
    \centering
    \begin{subfigure}[b]{0.49\columnwidth}
        \centering
        \includegraphics[width=\columnwidth]{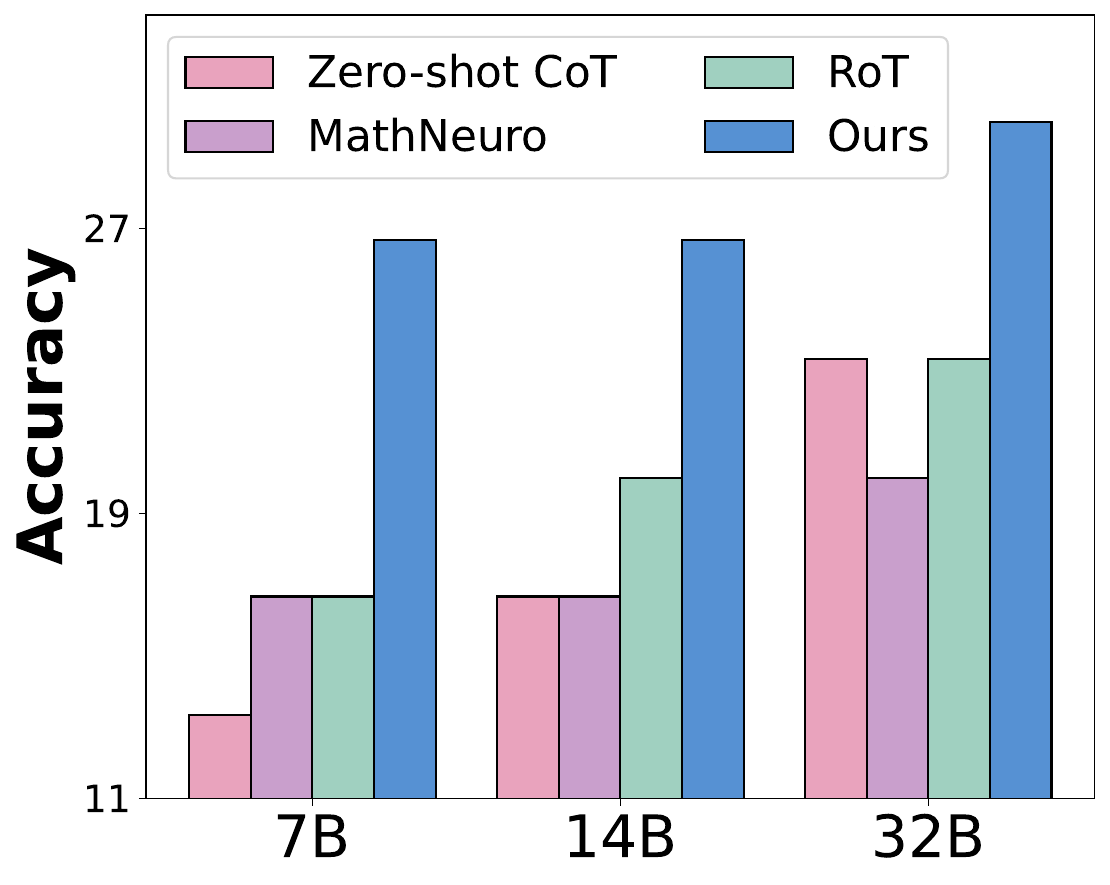}
        \caption{AIME}
        \label{fig:larger_aime}
    \end{subfigure}
    \begin{subfigure}[b]{0.49\columnwidth}
        \centering
        \includegraphics[width=\columnwidth]{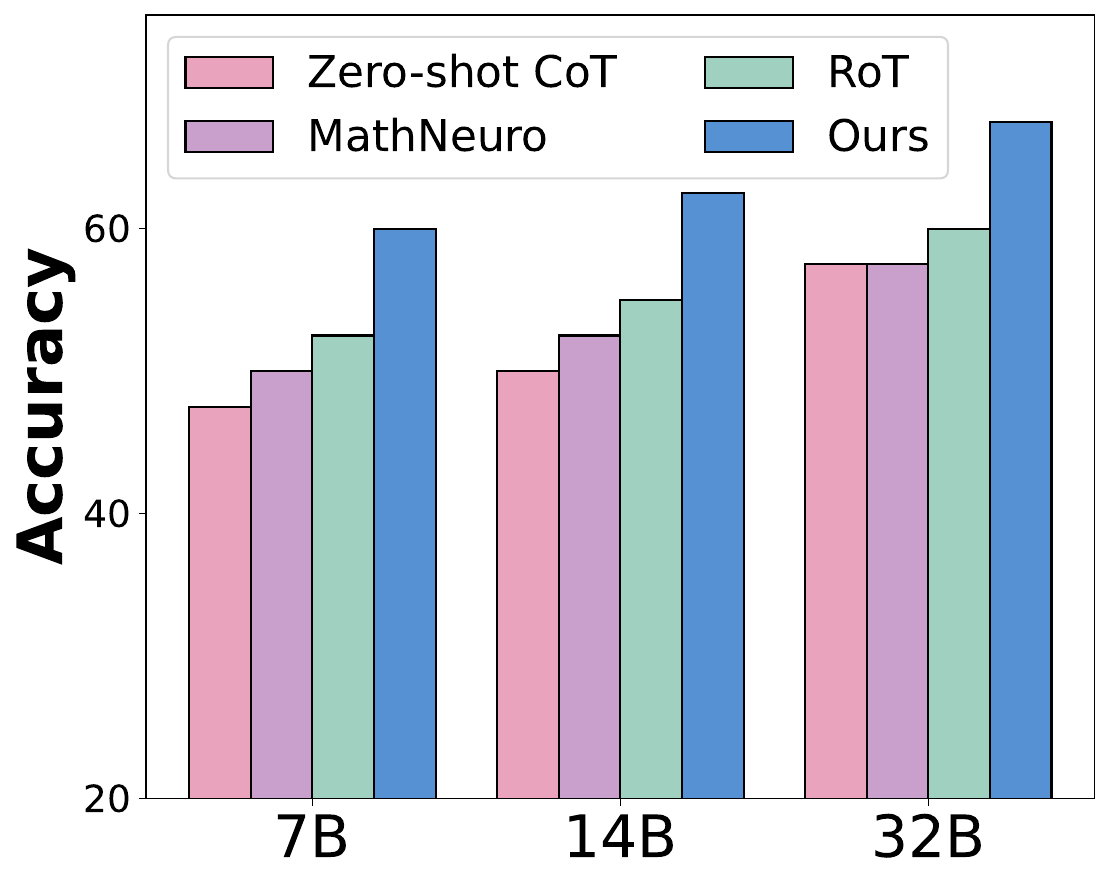}
        \caption{AMC}
        \label{fig:larger_amc}
    \end{subfigure}
    \caption{Performance comparison on AIME and AMC datasets using Qwen-series LLMs.}
\label{fig:larger_llms}
\end{figure}

In this part, we conduct experiments on Qwen-series LLMs using AIME and AMC datasets.
As illustrated in Figure~\ref{fig:larger_llms}, \OURS consistently outperforms all other baselines.
This further demonstrates the effectiveness of our proposed method.

\subsubsection{The Effect of the Representation Memory Size}

\begin{figure}[t]
    \centering
    \begin{subfigure}[b]{0.49\columnwidth}
        \centering
        \includegraphics[width=\columnwidth]{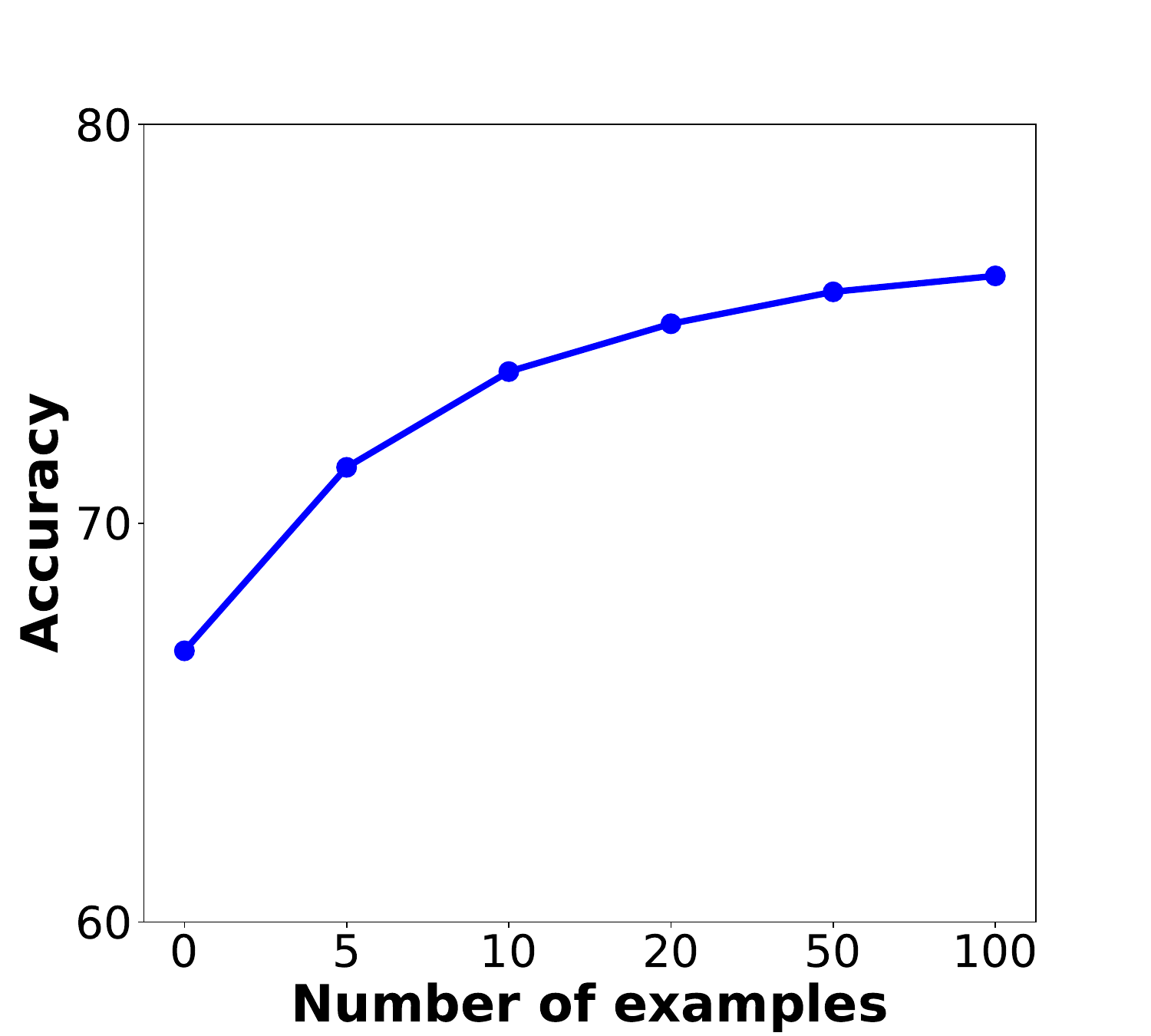}
        \caption{MATHOAI}
        \label{fig:in-domain-re_number}
    \end{subfigure}
    \begin{subfigure}[b]{0.49\columnwidth}
        \centering
        \includegraphics[width=\columnwidth]{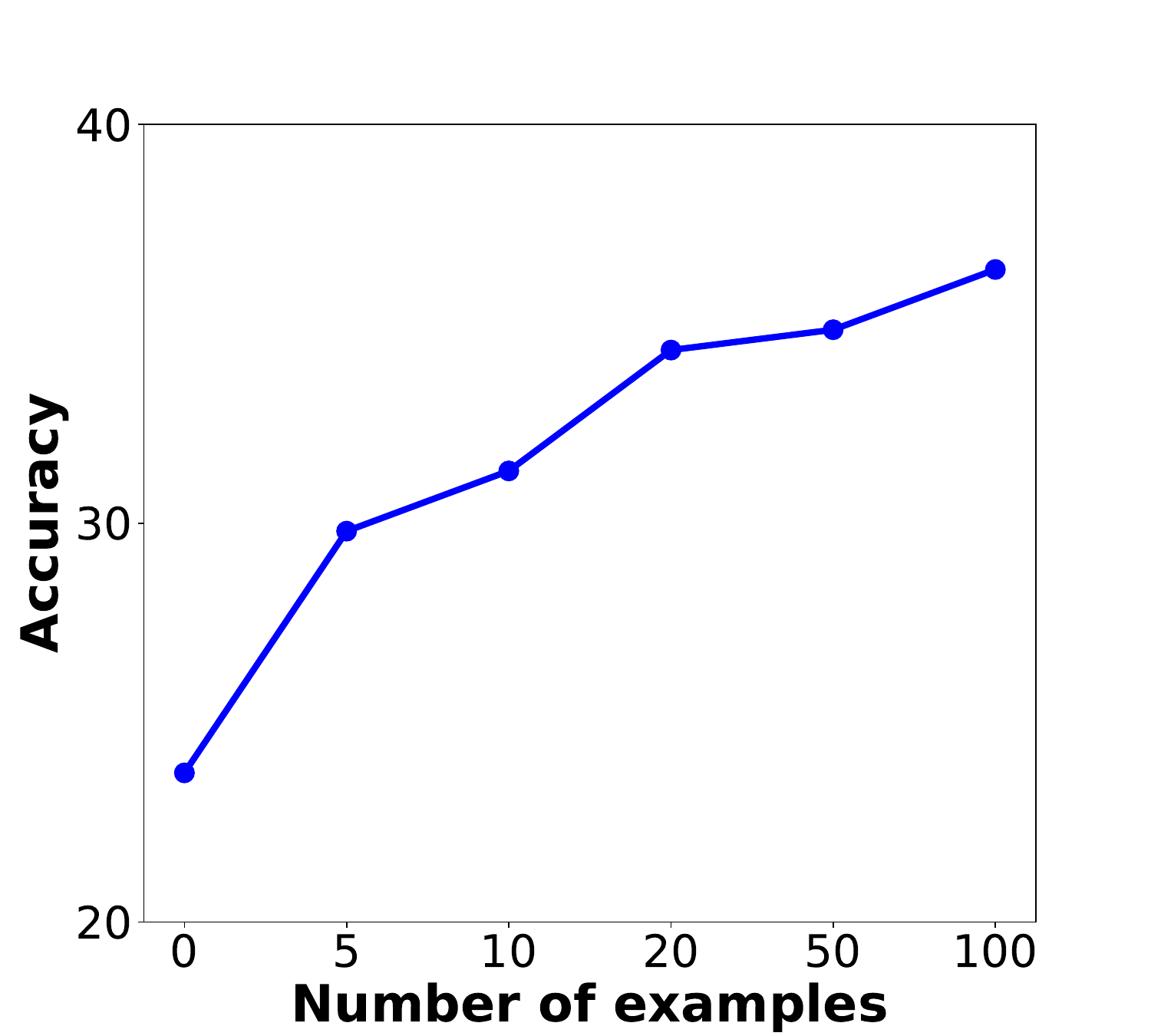}
        \caption{GPQA}
        \label{fig:cross-domain-re_number}
    \end{subfigure}
    \caption{Accuracy with increasing numbers of examples in the representation memory on the MATHOAI and GPQA datasets using Qwen2.5-7B-Instruct.}
\label{fig:re_number}
\end{figure}

In this part, we investigate the impact of scaling the representation memory size on \OURS.
The results are illustrated in~\ref{fig:re_number}.
As we can see, the performance consistently improves as the number of examples in the representation memory increases.
This improvement can be attributed to two main reasons.
On one hand, when computing {the average representation of the contrastive reasoning pattern representation}, using a larger number of demonstrations helps better isolate problem-specific information, resulting in more precise high-level long CoT pattern representations.
On the other hand, when extracting and injecting domain-specific features, the increased representation memory provides the model with access to more relevant question-aware domain-specific representations, enabling fine-grained refinement of the reasoning process in specific domains.

\subsubsection{The Impact of Layer Selection}

\begin{figure}[t]
    \centering
    \includegraphics[width=\columnwidth]{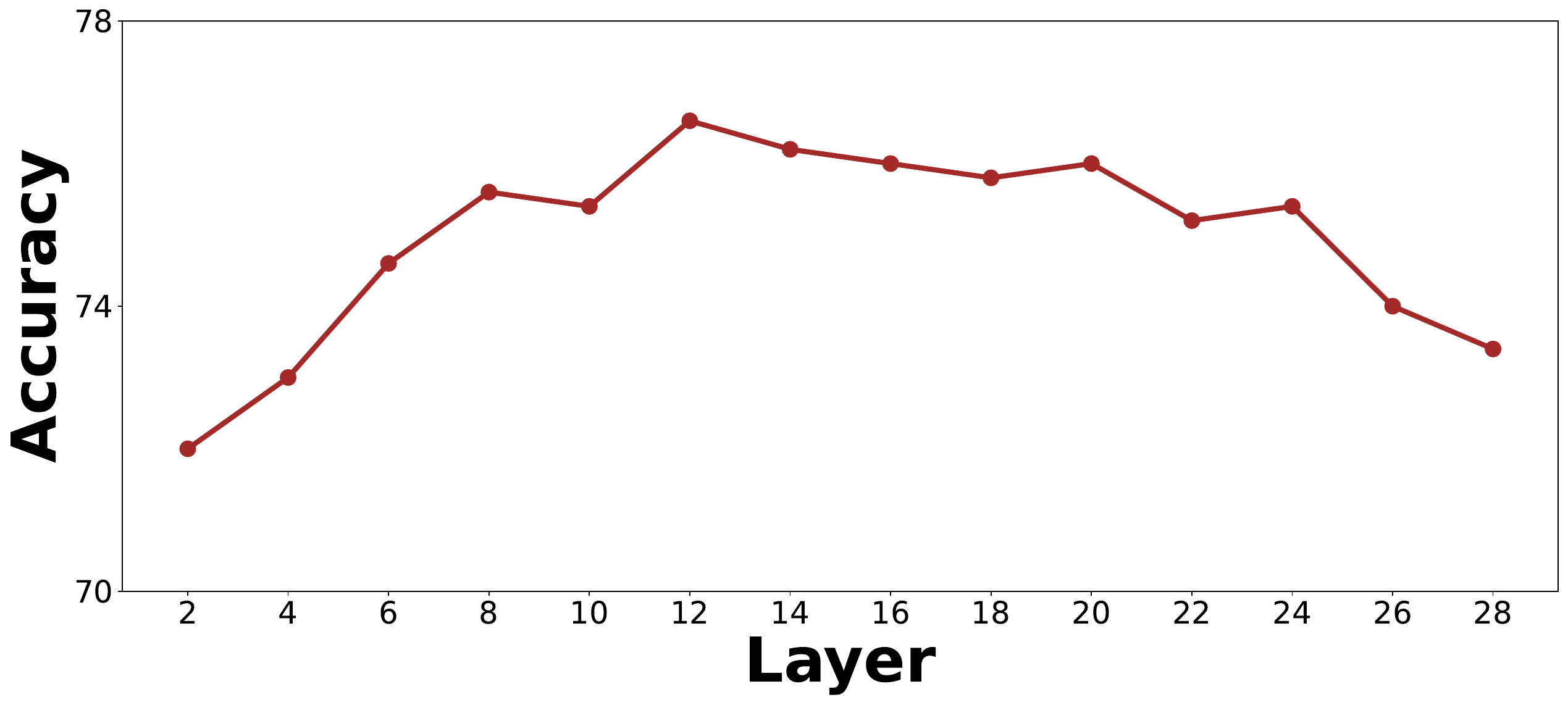}
    \caption{Performance comparison of the injection layer $L$ on the MATHOAI dataset using Qwen2.5-7B-Instruct.}
\label{fig:layer}
\end{figure}

In this section, we explore the impact of layer selection.
We conduct experiments across different layers on the MATHOAI dataset using Qwen2.5-7B-Instruct.
The results are presented in Figure~\ref{fig:layer}.
Our method exhibits a performance peak at the middle layer, with performance improving as the number of layers increases initially, but plateauing or declining in later layers.
Additionally, our method is not significantly affected by the layer selection, which demonstrates its robustness.

\subsubsection{Case Study}

In this part,  we demonstrate a specific example of how \OURS activates the long CoT reasoning capabilities of LLMs.
The case study example is illustrated in Figure~\ref{fig:case}.
Overall, compared to the zero-shot CoT method, \OURS can encourage LLMs to generate more intermediate reasoning steps, enabling them to engage in deliberate thinking.
Specifically, the zero-shot CoT method directly converts $y$ and $z$ into $x$, leading to errors in complex variable substitution and simplification, which disrupts the reasoning chain and results in calculation mistakes.
In contrast, \OURS introduces intermediate variables to simplify the reasoning process and structures the problem-solving approach in a step-by-step manner.
This approach helps maintain logical consistency throughout the reasoning process, significantly activating the long CoT reasoning capabilities of LLMs on complex reasoning problems.
\section{Conclusion}
\label{sec-conclusion}

In this work, we conduct an empirical analysis for the mechanism of long CoT reasoning from the perspective of representation.
Our findings reveal that long CoT reasoning appears to be a general capability potentially encoded in LLMs.
Inspired by this, we propose a novel training-free method based on representation engineering, which can effectively and efficiently unleash the general long CoT reasoning capabilities of LLMs.
Overall, our work provides a deeper understanding of long CoT reasoning, paving the way for transparent and interpretable slow-thinking reasoning models.

\section{Limitations}
\label{sec-limitations}

One limitation of our work is that our method requires access to the internal representations of the model, making it infeasible for closed-source LLMs.
In addition, due to the constraints of our cost and resources, we only conduct experiments on representative tasks and LLMs.

\section*{Acknowledgements}
This work was partially supported by National Natural Science Foundation of China under Grant No. 92470205 and 62222215, Beijing Municipal Science and Technology Project under Grant No. Z231100010323009, Beijing Natural Science Foundation under Grant No. L233008, and Ant Group.
Xin Zhao is the corresponding author.

\bibliography{newbib}

\newpage

\appendix

\section{Quantitative Analysis of Vanilla and Long CoT Representations.}
\label{app:qa-CoT}

After discovering the distinct distributions of vanilla and long CoTs within LLMs, we further conduct a quantitative analysis of their representations.
Specifically, we employ matrix-based entropy~\cite{Giraldo-2015-Measures-arXiv,Wei-2024-Large-arXiv} to investigate the information content across different layers for both CoTs.
Given the representations of $n$ samples $Z \in \mathbb{R}^{n*d}$, the matrix-based entropy $S_{\alpha}(\mathbf{Z})$ quantifies the diversity of features within the representations, as defined by the following equations:
\begin{align}
    \mathbf{K} &= \mathbf{Z}\mathbf{Z}^{\top}, \\
    S_{\alpha}(\mathbf{Z}) = \frac{1}{1 - \alpha} \log & \left( \sum_{i=1}^{r} \left( \frac{\lambda_i(\mathbf{K})}{\mathrm{tr}(\mathbf{K})} \right)^{\alpha} \right),
\end{align}

where $\mathbf{K}$ is the Gram matrix of the representation $\mathbf{Z}$, $\lambda_i(\mathbf{K})$ represents the nonnegative eigenvalues of $\mathbf{K}$, and $r= \text{rank}(\mathbf{K})\leq \min(d,n)$.
Following ~\citet{Skean-2025-Layer-arXiv}, we set $\alpha=1$ for simplicity.

\begin{figure}[t]
    \centering
    \begin{subfigure}[b]{0.49\columnwidth}
        \centering
        \includegraphics[width=\columnwidth]{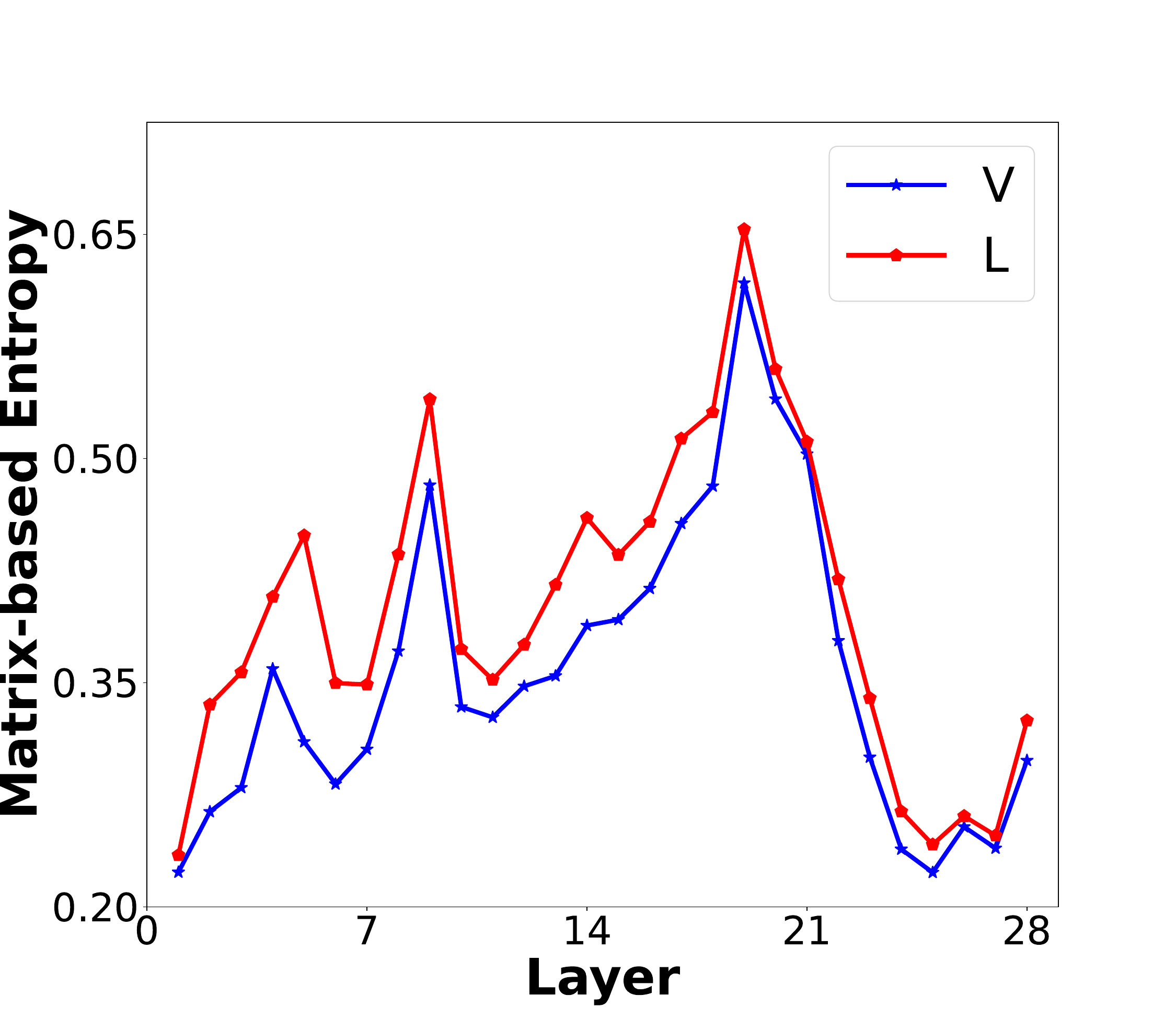}
        \caption{Qwen2.5-7B-Instruct}
        \label{fig:qwen-entropy}
    \end{subfigure}
    \begin{subfigure}[b]{0.49\columnwidth}
        \centering
        \includegraphics[width=\columnwidth]{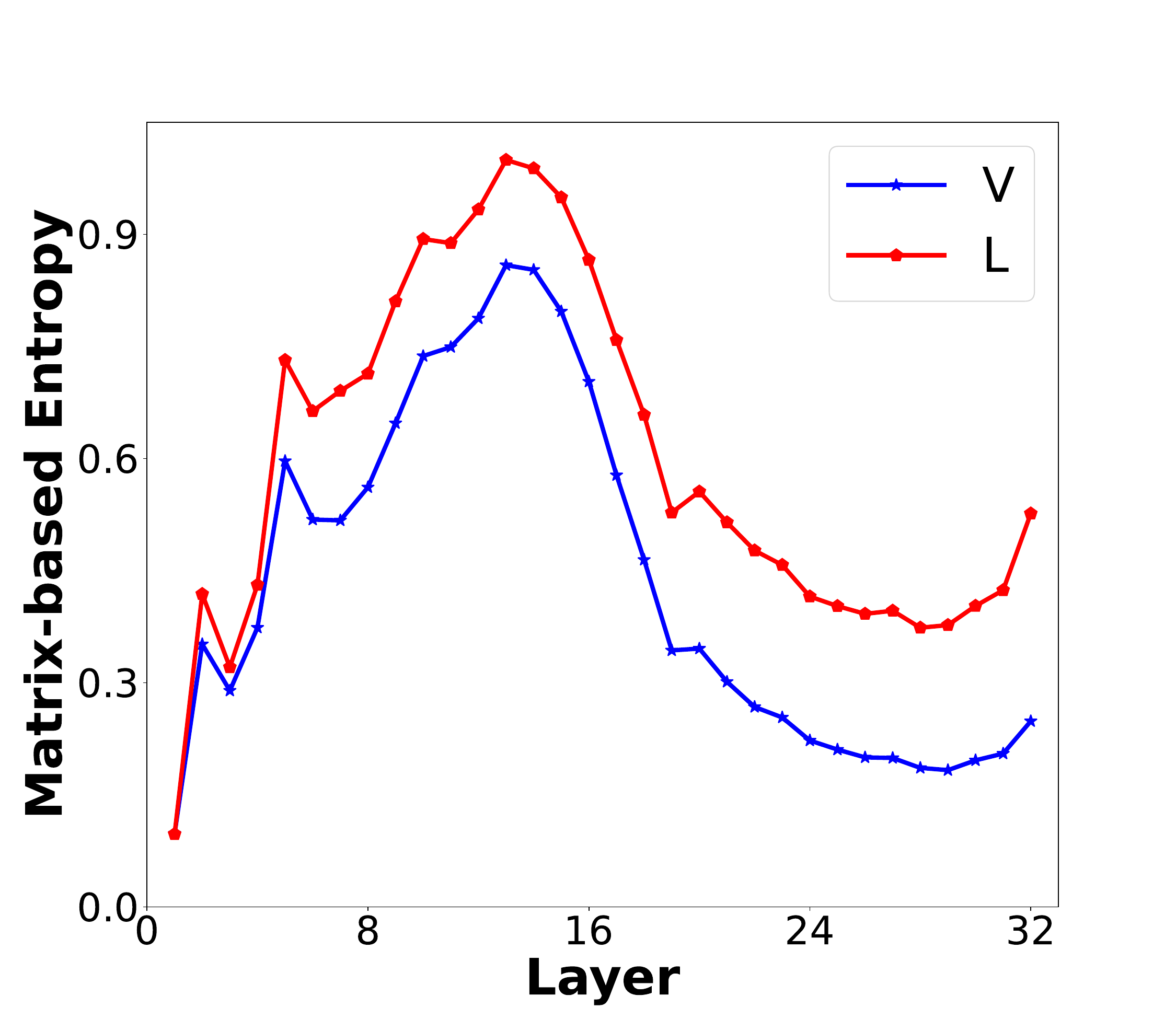}
        \caption{Llama3.1-8B-Instruct}
        \label{fig:llama-entropy}
    \end{subfigure}
    \caption{Matrix-based entropy across all layer in Qwen2.5-7B-Instruct and Llama3.1-8b-Instruct. ``V'' and ``L'' denote the vanilla and long CoT, respectively.}
\label{fig:entropy}
\end{figure}

The matrix-based entropy metrics for vanilla and long CoTs representations across different layers in Qwen2.5-7B-Instruct and Llama3.1-8B-Instruct are illustrated in Figure~\ref{fig:entropy}.
We observe that the matrix-based entropy of long CoT is consistently higher than that of vanilla CoT, indicating that long CoT contains more diverse and less redundant features within the latent space.
Additionally, we find that the entropy in the middle layers of the model is higher than in the final layer in both CoTs.
This suggests that the middle layers are better at extracting diverse and complex features~\cite{RAG-mechanism}, exhibiting powerful capabilities in reasoning tasks~\cite{ElNouby-2024-Scalable-ICML,Fan-2024-Not-arXiv}.
\section{Statistics of vanilla and long CoT Examples}
\label{app:dataset-stastics}

\begin{table}[t]
    \centering
    \resizebox{\columnwidth}{!}{
    \begin{tabular}{l|cccc}
        \toprule
        Domain & Math & Physics & Chemistry & Biology \\
        \midrule
        Average Tokens of Vanilla CoT  & 400.98  & 365.45  & 356.59  & 347.73 \\
        Average Tokens of Long CoT     & 2628.46 & 2094.35 & 1832.86 & 1607.29 \\
        \bottomrule
    \end{tabular}
    }
    \caption{Statistics of the vanilla and long CoT examples.}
    \label{tab:examples_thought}
\end{table}

In this paper, we leverage open-source data from STILL-2~\cite{STILL2}, a high-quality dataset consisting of vanilla and long CoTs distilled from DeepSeek-R1-Lite-Preview~\cite{deepseek-r1}.
From this dataset, we randomly select 100 examples from the math, physics, chemistry, and biology domains, respectively.
Table~\ref{tab:examples_thought} provides detailed statistics of the vanilla and long CoT examples.
\section{Detailed Visualization of Short-form and Long-form Thought Representations}
\label{app:all-representations}

\begin{figure*}[t]
    \centering
    \includegraphics[width=\linewidth]{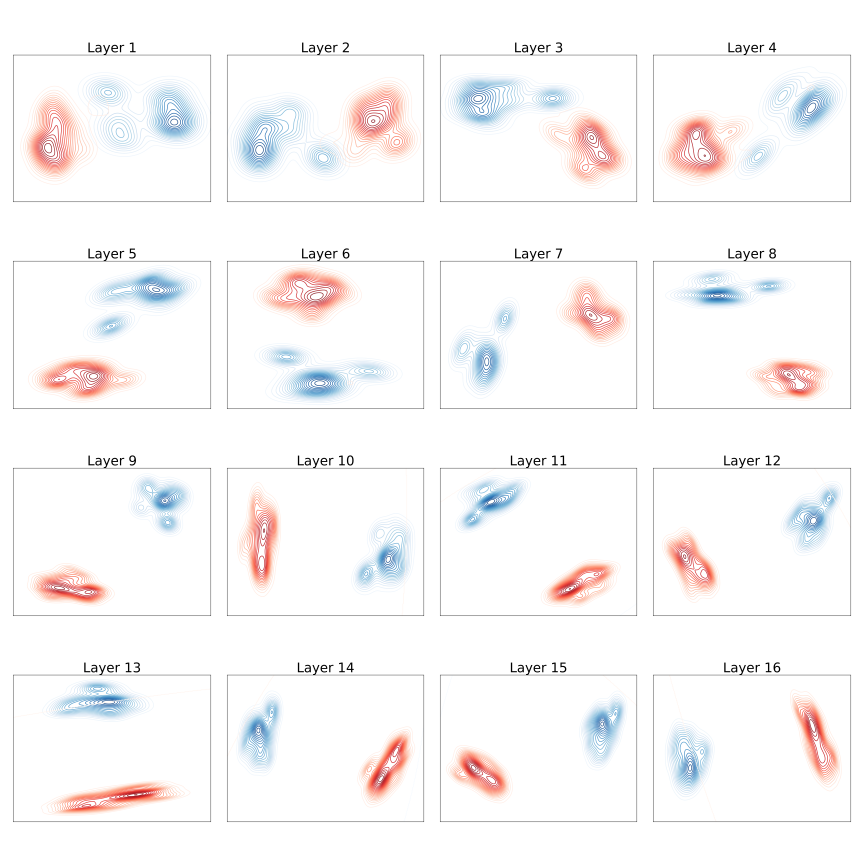}
    \caption{t-SNE plot of Qwen2.5-7B-Instruct’s representations for vanilla~(blue) and long CoTs~(red) across 1-16 layers.}
\label{fig:representations_qwen_all1}
\end{figure*}

\begin{figure*}[t]
    \centering
    \includegraphics[width=\linewidth]{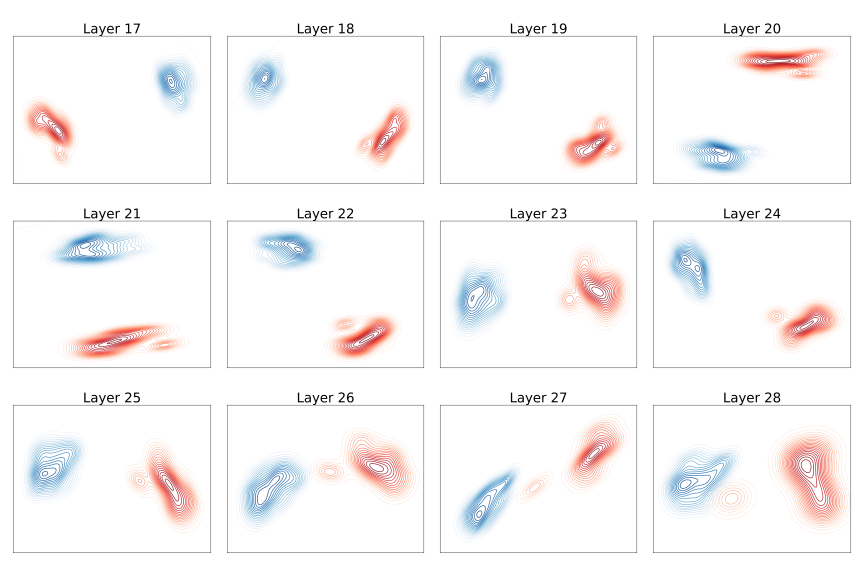}
    \caption{t-SNE plot of Qwen2.5-7B-Instruct’s representations for vanilla~(blue) and long CoTs~(red) across 17-28 layers.}
\label{fig:representations_qwen_all2}
\end{figure*}

\begin{figure*}[t]
    \centering
    \includegraphics[width=\linewidth]{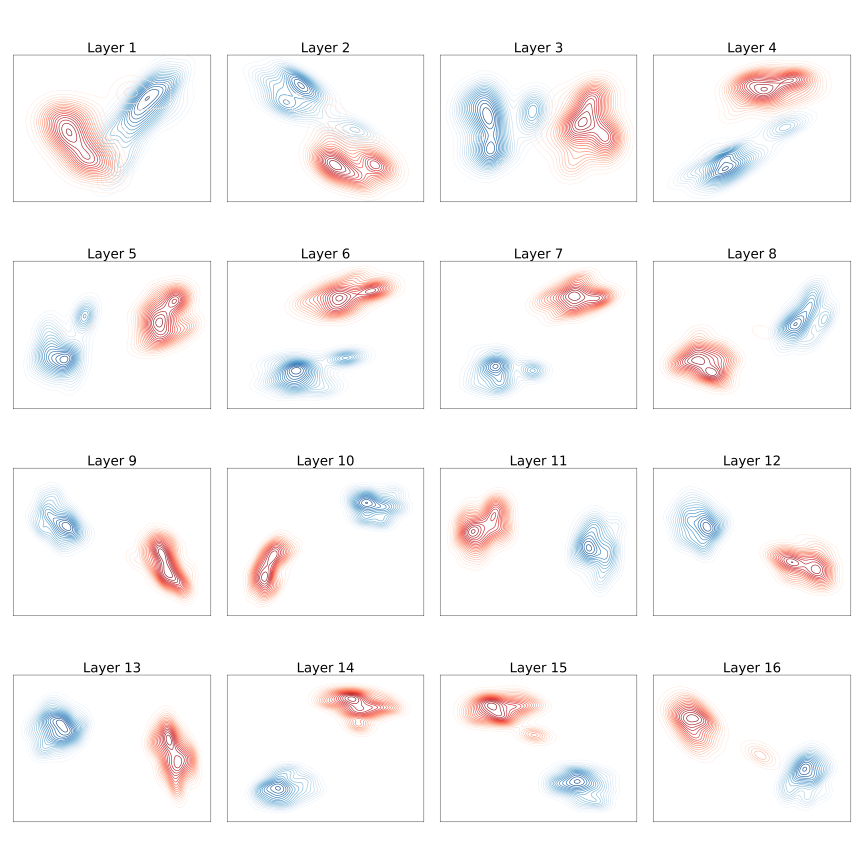}
    \caption{t-SNE plot of Llama3.1-8B-Instruct’s representations for vanilla~(blue) and long CoTs~(red) across 1-16 layers.}
\label{fig:representations_llama_all1}
\end{figure*}

\begin{figure*}[t]
    \centering
    \includegraphics[width=\linewidth]{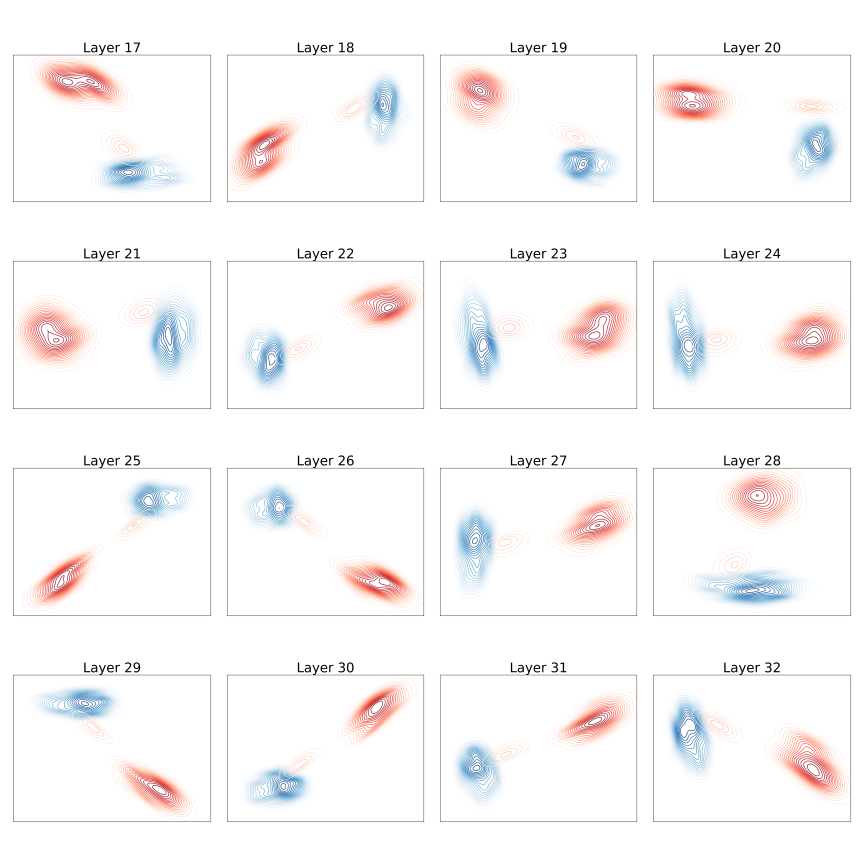}
    \caption{t-SNE plot of Llama3.1-8B-Instruct’s representations for vanilla~(blue) and long CoTs~(red) across 17-32 layers.}
\label{fig:representations_llama_all2}
\end{figure*}

In this section, we present the t-SNE visualizations of Qwen2.5-7B-Instruct's and Llama3.1-8B-Instruct's representations for vanilla and long CoTs across all layers, as illustrated in Figure~\ref{fig:representations_qwen_all1}, ~\ref{fig:representations_qwen_all2}, ~\ref{fig:representations_llama_all1}, and ~\ref{fig:representations_llama_all2}, respectively.

\section{Detailed Visualization of Representations across Different Domains}
\label{app:all-representations-domain}

\begin{figure*}[t]
    \centering
    \begin{subfigure}[b]{0.326\textwidth}
        \centering
        \includegraphics[width=\textwidth]{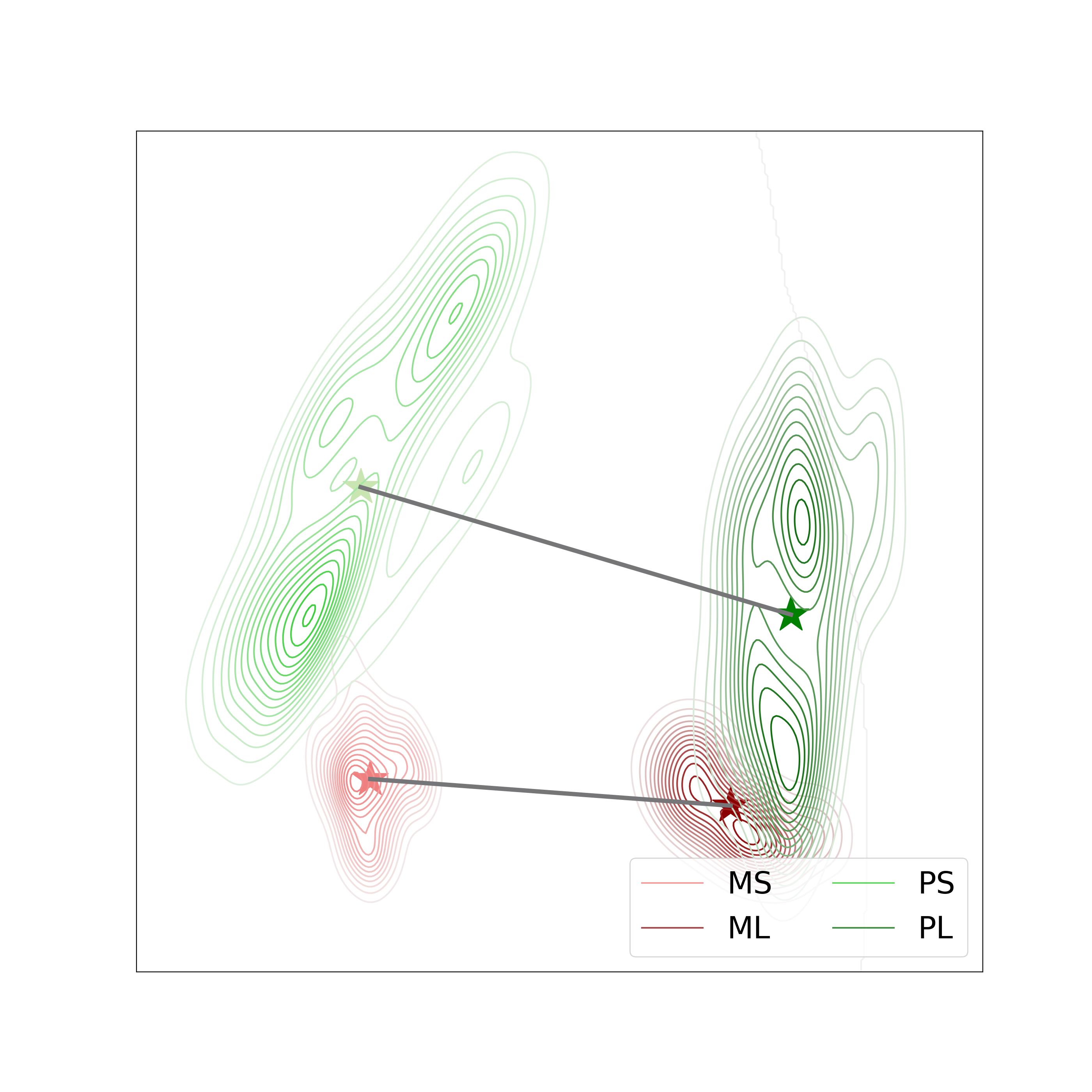}
        \caption{Qwen~(Math and Physics)}
        \label{fig:q-math-phisics}
    \end{subfigure}
    \begin{subfigure}[b]{0.326\textwidth}
        \centering
        \includegraphics[width=\textwidth]{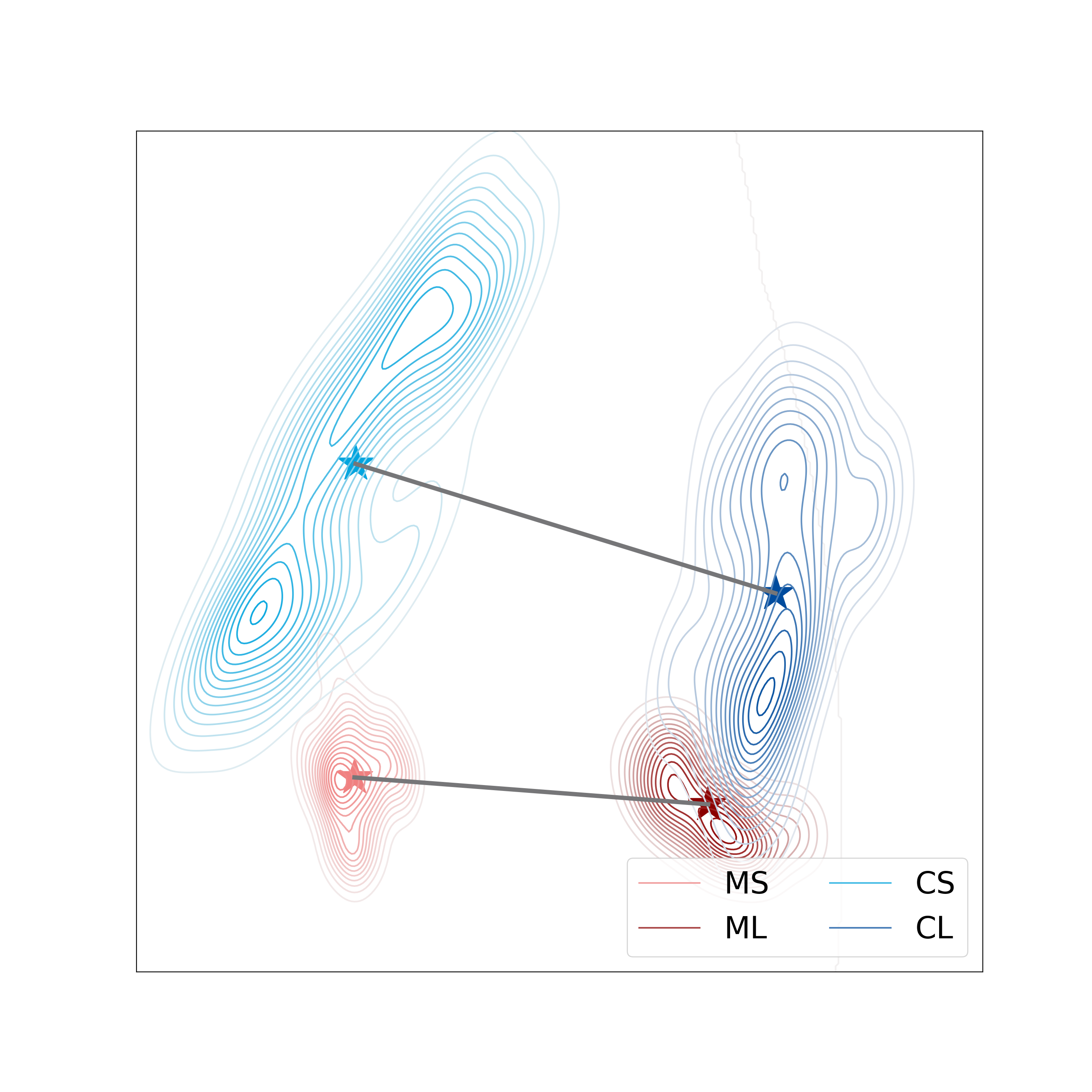}
        \caption{Qwen~(Math and Chemistry)}
        \label{fig:q-math-chemistry}
    \end{subfigure}
    \begin{subfigure}[b]{0.326\textwidth}
        \centering
        \includegraphics[width=\textwidth]{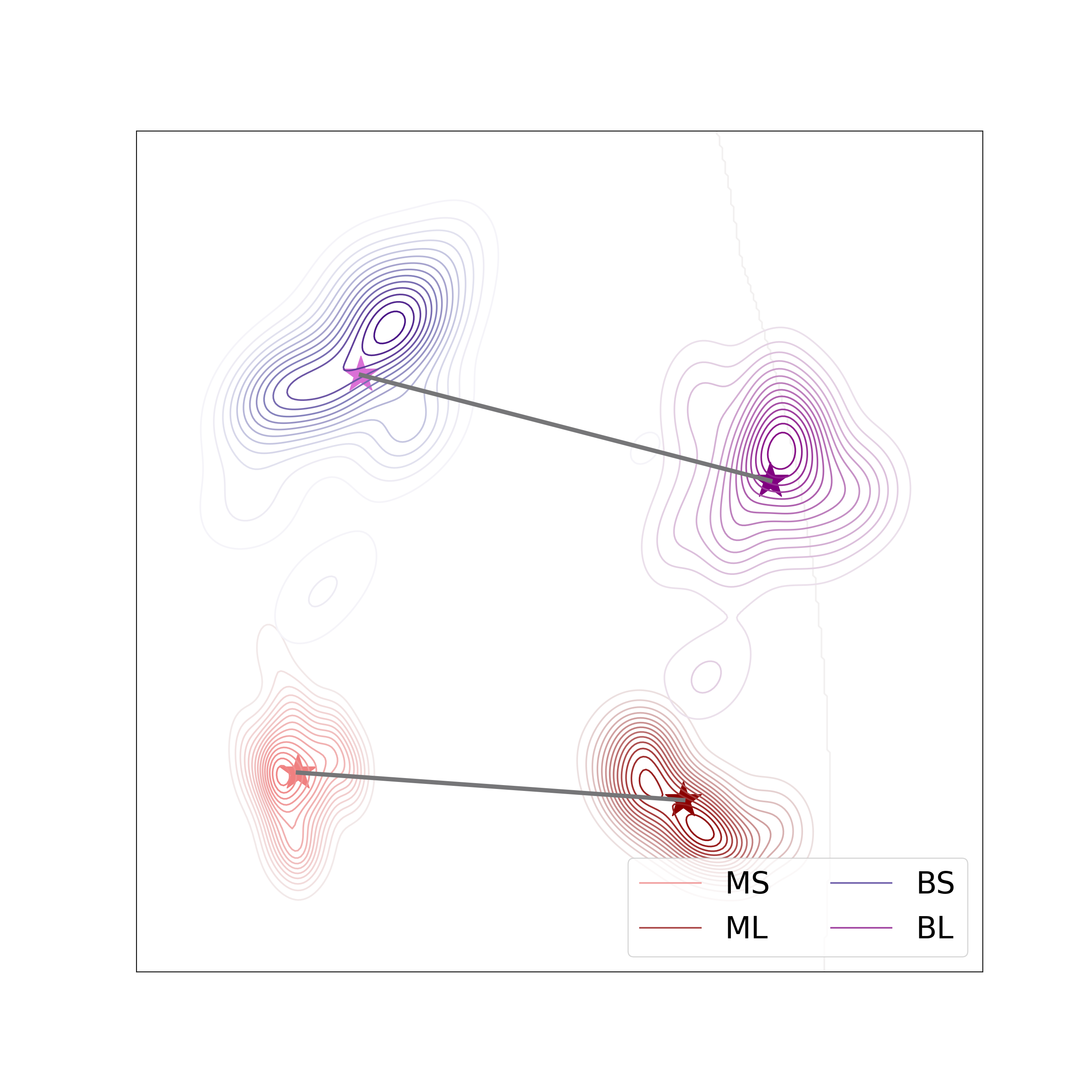}
        \caption{Qwen~(Math and Biology)}
        \label{fig:q-math-biology}
    \end{subfigure}

    \centering
    \begin{subfigure}[b]{0.326\textwidth}
        \centering
        \includegraphics[width=\textwidth]{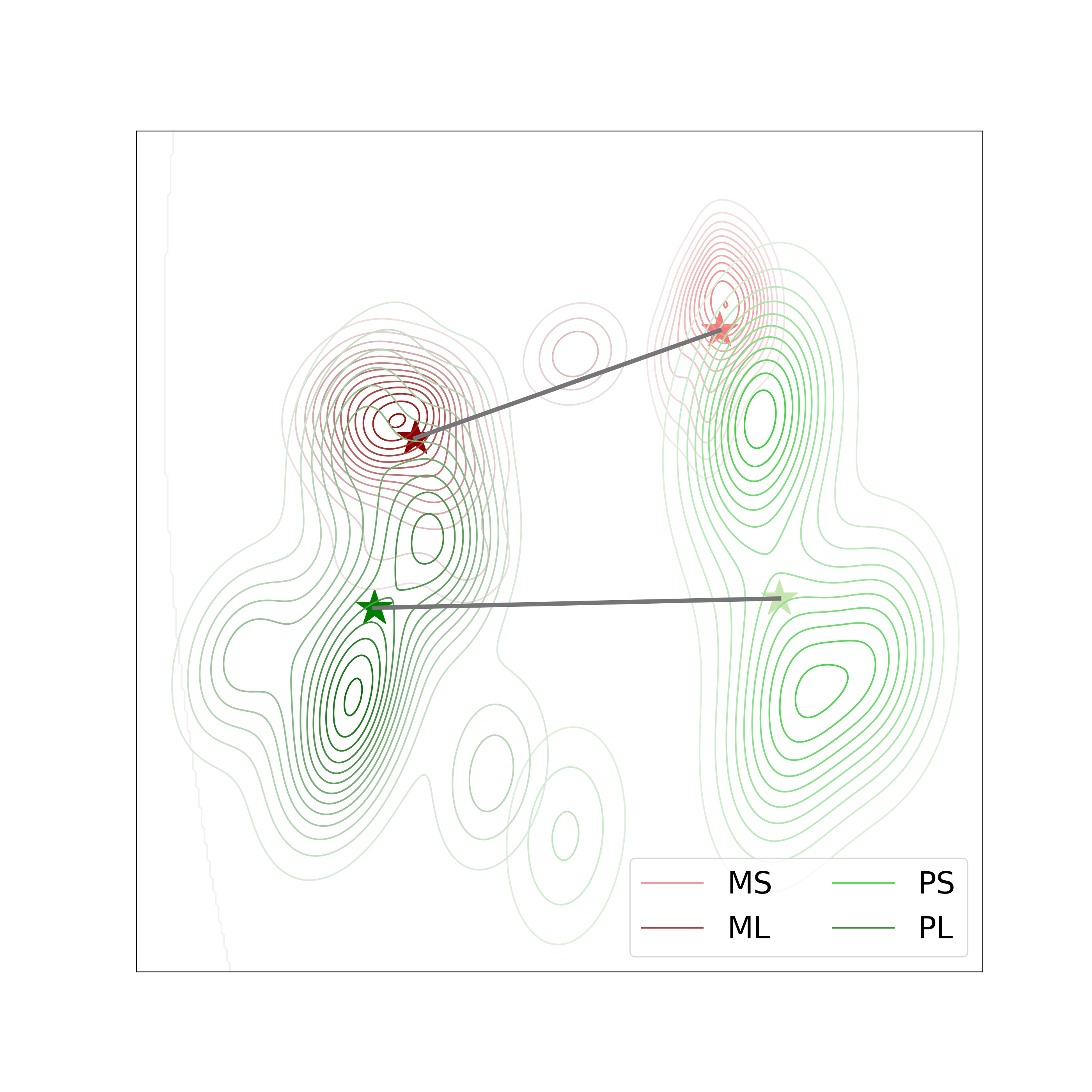}
        \caption{Llama~(Math and Physics)}
        \label{fig:l-math-phisics}
    \end{subfigure}
    \begin{subfigure}[b]{0.326\textwidth}
        \centering
        \includegraphics[width=\textwidth]{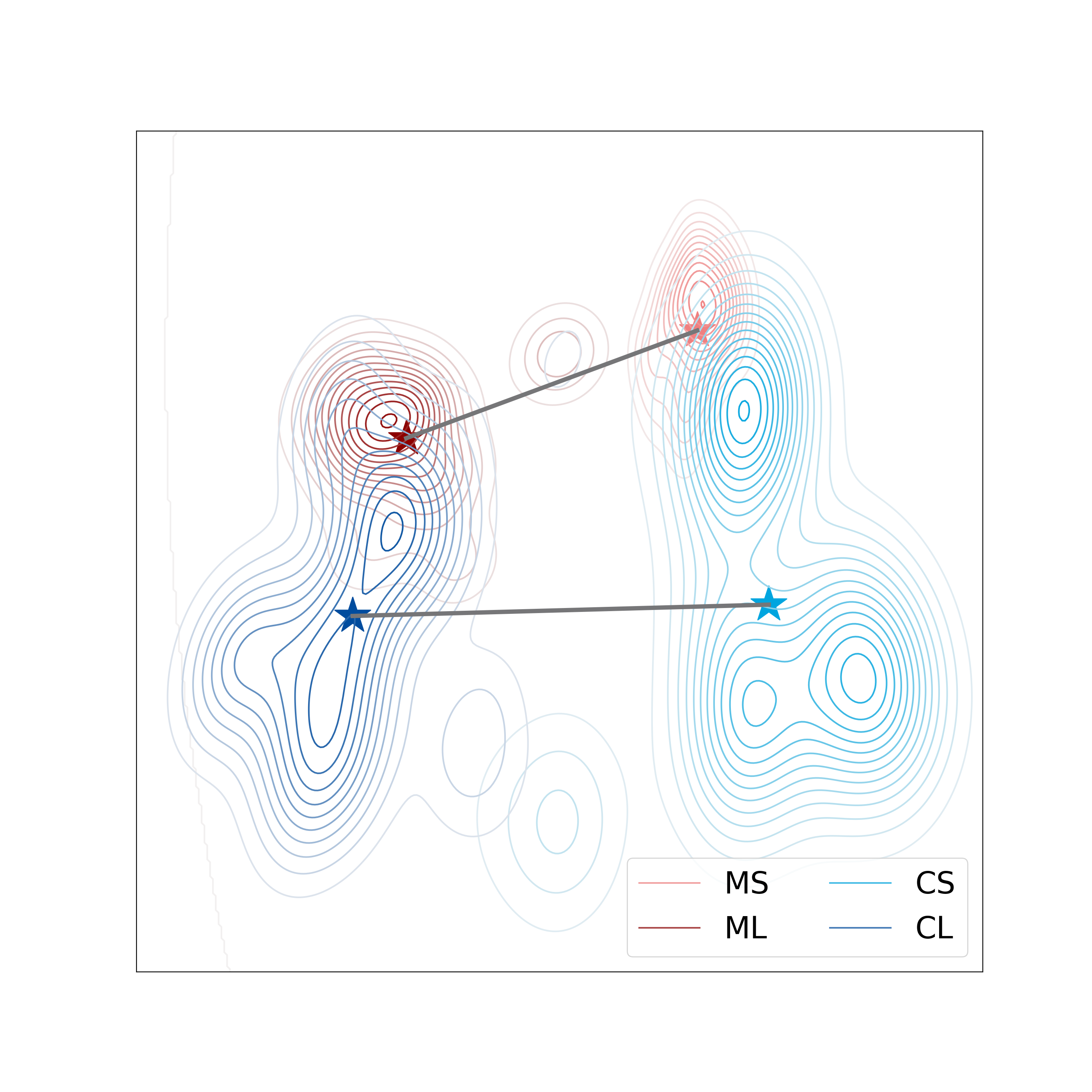}
        \caption{Llama~(Math and Chemistry)}
        \label{fig:l-math-chemistry}
    \end{subfigure}
    \begin{subfigure}[b]{0.326\textwidth}
        \centering
        \includegraphics[width=\textwidth]{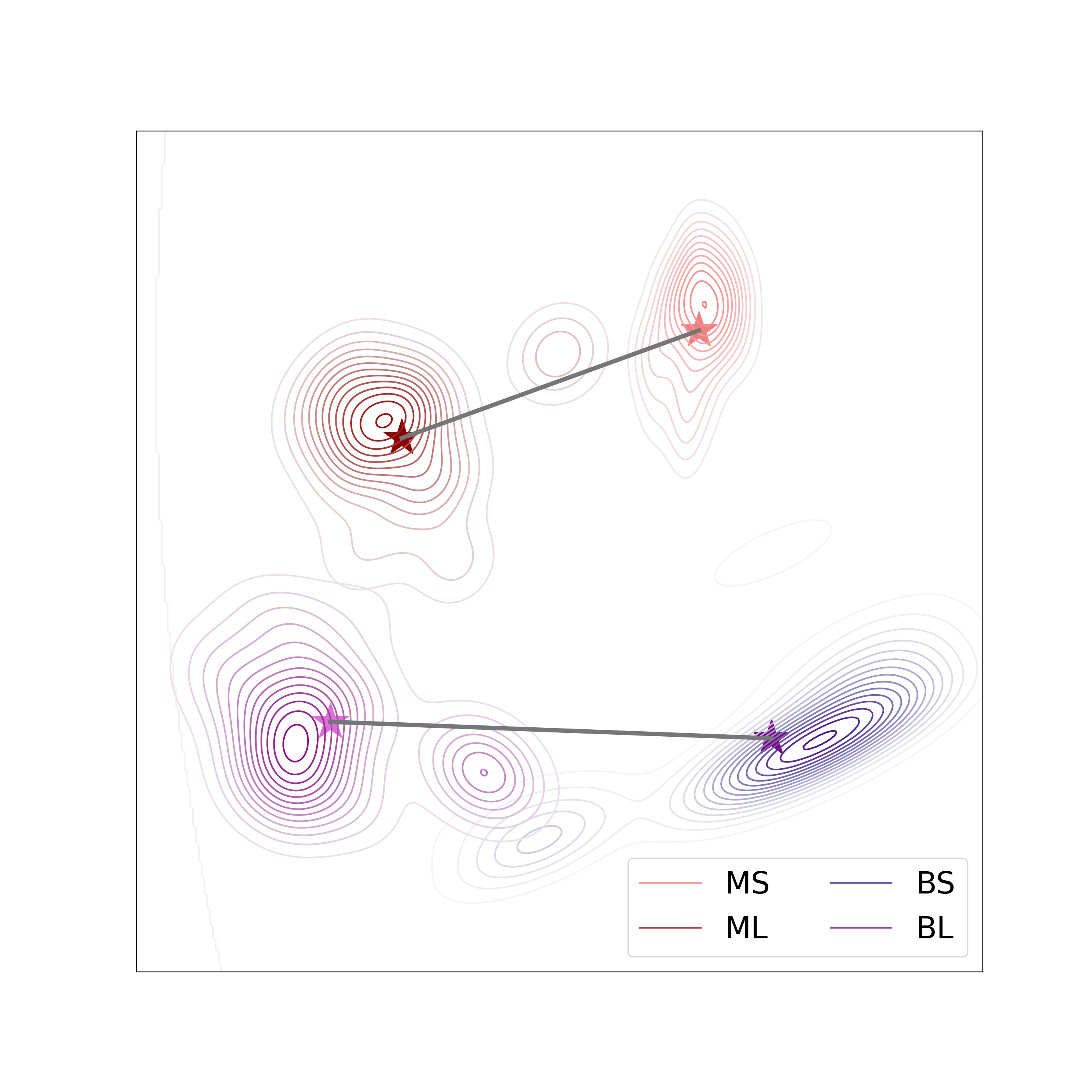}
        \caption{Llama~(Math and Biology)}
        \label{fig:l-math-biology}
    \end{subfigure}
    
    \caption{t-SNE plot of representations from Qwen2.5-7B-Instruct and Llama3.1-8B-Instruct for vanilla and long CoTs across math and other domains~(\ie physics, chemistry and biology). ``MS'', ``PS'', ``CS'', and ``BS'' denote the vanilla CoT on the math, physics, chemistry, and biology domains, respectively. ``ML'', ``PL'', ``CL'', and ``BL'' denote the long CoT on these domains.}
\label{fig:domain-re}
\end{figure*}

In this section, we present detailed visualizations of vanilla and long CoT representations in the middle layers of LLMs across math and other domains~(\ie physics, chemistry, and biology).
The results are shown in Figure~\ref{fig:domain-re}.
\section{Detailed Description of Baselines.}
\label{app:baselines}

In this part, we provide detailed descriptions of all the baselines used in our experiments.
These include prompting-based approaches~(\ie Zero-shot CoT, Few-shot CoT, and BoostStep~\cite{arXiv-Zhang-BoostStep}), neuron activation method~(\ie MathNeuro~\cite{MathNeuro}), representation engineering method~(\ie RoT~\cite{RoT}), and supervised fine-tuning method.

$\bullet$ \textbf{\underline{Zero-shot CoT}}:
The model generates answers directly using only the problem and a CoT prompt~(\ie Answer the following question step by step and put the final answer in \textbackslash \textbackslash boxed\{\}) as input, without any additional demonstrations.

$\bullet$ \textbf{\underline{Few-shot CoT}}:
The model makes predictions with long CoT examples and a CoT prompt.

$\bullet$ \textbf{\underline{BoostStep}}~\cite{arXiv-Zhang-BoostStep}:
This method guides the model to perform the reasoning process incrementally and provides similar step-level examples at each reasoning step.

$\bullet$ \textbf{\underline{MathNeuro}}~\cite{MathNeuro}
This method leverages weights and activations from the forward pass to identify and isolate specific parameters associated with reasoning capabilities, and enhances the model's reasoning performance through pruning and scaling of these parameters.

$\bullet$ \textbf{\underline{RoT}}~\cite{RoT}:
This method extracts contrastive representations based on whether a CoT prompt or a non-CoT prompt is included in the input, and then injects them into the model's latent space.

$\bullet$ \textbf{\underline{SFT}}:
This method employs a supervised fine-tuning method on 100 long-form thought samples from each of four domains and performs zero-shot CoT during inference.

\end{document}